\pdfoutput=1

\documentclass[11pt,a4paper,oneside]{article}

\usepackage[T1]{fontenc}
\usepackage[utf8]{inputenc}
\usepackage{lmodern}

\usepackage[left=3.5cm,right=2.0cm,top=2.5cm,bottom=2.5cm]{geometry}
\usepackage{setspace}
\usepackage[british]{babel}
\usepackage{csquotes}
\usepackage{amsmath,amssymb}
\usepackage{graphicx}
\graphicspath{{figures/}{figs/}}
\usepackage{booktabs}
\usepackage{xcolor}
\usepackage[framemethod=TikZ]{mdframed}
\usetikzlibrary{arrows.meta,positioning,fit,backgrounds,decorations.pathreplacing}

\definecolor{ag}{HTML}{D98B10}
\definecolor{tp}{HTML}{E8C97A}
\definecolor{en}{HTML}{E8622A}
\definecolor{st}{HTML}{E0BA83}

\usepackage[authoryear,round]{natbib}
\PassOptionsToPackage{starttoc}{rerunfilecheck}
\usepackage[backref=page]{hyperref}
\usepackage{cleveref}

\newcommand{\newsection}[1]{\clearpage\section{#1}}

\title{Cooperative Multi-Agent Reinforcement Learning\\for Market Making in Prediction Markets}
\author{}
\date{}

\begin{document}

\pagenumbering{roman}

\newgeometry{left=2.5cm,right=2.5cm,top=2.5cm,bottom=2.5cm}

\begin{titlepage}
\centering
\thispagestyle{empty}
\setstretch{1.0}

\vspace*{0.45cm}

{\fontsize{23}{27}\selectfont\bfseries
Cooperative Multi-Agent Market Making
on Prediction-Market Order Books\par}

\vspace{2.75cm}

\includegraphics[
    width=0.34\textwidth,
    height=4.4cm,
    keepaspectratio
]{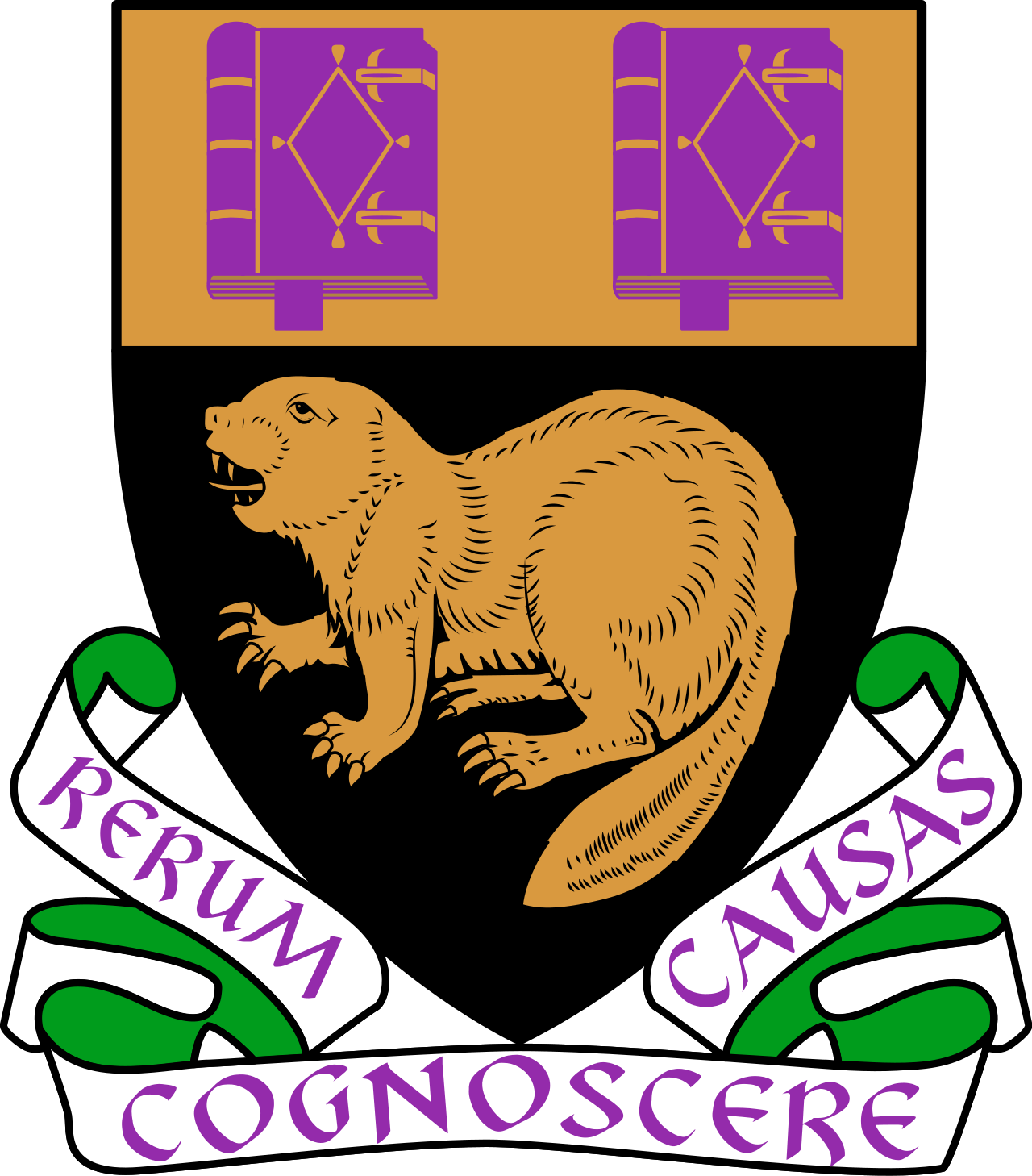}

\vspace{2.45cm}

{\fontsize{17}{21}\selectfont
Skander Lejmi\par}

\vspace{0.30cm}

{\fontsize{15}{19}\selectfont
Department of Statistics\par}

\vspace{0.22cm}

{\fontsize{15}{19}\selectfont
London School of Economics and Political Science\par}

\vspace{1.75cm}

{\fontsize{13}{18}\selectfont
A thesis submitted for the degree of\par}

\vspace{0.32cm}

{\fontsize{14}{19}\selectfont\itshape
MSc Statistics (Research)\par}

\vspace{0.85cm}

{\fontsize{14}{19}\selectfont
Lent Term, 2026\par}

\vspace{1.55cm}

{\fontsize{13}{18}\selectfont
Supervised by Professor Tengyao Wang\par}

\vfill

\end{titlepage}

\restoregeometry

\thispagestyle{empty}

\clearpage
\section*{Abstract}
\addcontentsline{toc}{section}{Abstract}

A market maker's objective is the provision of liquidity. To do so, it quotes prices to buy and to sell a given asset, earning the gap between
them (spread), and carrying the risk of the position it builds up. We study
market makers trained with multi-agent reinforcement learning on prediction markets, specifically Polymarket. 
These exchanges are characterized by their relatively thin order books whose binary contracts pay exactly 0 or 1 at
a known resolution date.

Three challenges underpin this research. Firstly, a maker faces adverse selection: the trades it receives come
disproportionately from better-informed traders, so a policy trained on real data may learn to stop quoting. The consequences of a given quote may arrive many steps after the action.
And prediction markets L2 data does not cover individual orders that a sophisticated simulator may require.

We address these challenges by building a pipeline in order to reconstruct LOBSTER data from Polymarket. 
This data pipeline subsequently allows for the training of various market-making agents in the JaxMARL-HFT environment on 417 real markets. 
We focus on improving the market making capabilities of the default configuration in JaxMARL-HFT, through enriching the observation space with the 
book and the recent flow. We additionally make contributions with regard to reward shaping, and crucially propose a novel multi-agent 
decomposition of the market-making agent into a quoter and a risk manager, enabling a more sophisticated treatment of the inventory risk. 
We find that decomposition alone performs poorly, while reward designs intended to mitigate multi-agent credit-assignment difficulties substantially improve the resulting desk relative to the unshaped decomposition in our experiments.

Our redesigned single-agent market maker quotes more frequently and incurs smaller held-out losses than the JaxMARL-HFT baseline retrained on the same data; the contributions of its jointly changed observations, actions and risk constraints are not separately identified.
Our multi-agent decomposition of the seat presents our central result: careful design of the reward split gives
us our best result. Our models are scored on realised end-of-episode value on the held-out set of markets. None establishes positive average portfolio value in periodic held-out evaluation, and each configuration was trained with one seed.

The environment, data and protocol are the foundation for makers that
quote many linked markets at once.

\clearpage

\tableofcontents
\clearpage
\listoffigures
\clearpage
\listoftables
\clearpage

\pagenumbering{arabic}
\setcounter{page}{1}

\newsection{Introduction}

High-frequency trading in financial markets is fundamentally a problem of
sequential decision making under uncertainty. Whether the objective is to
capture some market-orthogonal return (alpha), provide liquidity to the
markets (market making) or allocate capital across different markets based on
one's risk tolerance (portfolio optimisation), permeating all three is a
central challenge: prices form endogenously and evolve continuously as a
function of the decisions of millions of market participants, each acting on
private (and often partial) information. It is in this precise sense that the
market price is often referred to as an information aggregator.

A market maker supplies liquidity to markets by continuously quoting prices
at which it would be willing to buy (bid) and sell (ask) a given asset. In
doing so, it seeks to maximise its net profits while simultaneously managing
the inventory risk of its filled positions. Given the dynamic nature of
financial markets this is a non-trivial challenge, as the price of the asset
may move against whatever the maker is holding. With the advent of electronic
trading, the move to automate market making has produced a range of
approaches, from closed-form stochastic control to learned policies; we
review them in \cref{sec:related}.

Information aggregation is prominently the value proposition of recently
established prediction markets: exchanges offering binary, fixed-term
contracts on a far wider range of events than the financial system covers.
These venues have attracted academic attention in large part due to the
novelty of their contractual offering; the term \emph{combinatorial
arbitrage} has emerged for configurations in which two or more contracts are
logically constrained by one another, and participants appear to exploit such
structures to extract substantial edge \citep{saguillo2025arbitrage}.
Liquidity provision on these venues is independently quite interesting: given
the relative thinness of their books, a competent maker is a service the exchange could benefit from (as evidenced
by their recent \$1 million contribution to rebates for market making on the very markets we are considering \citep{polymarket2026rewards}).
This makes market making the natural first problem to study on these novel exchanges. They also publish full historical L2-orderbook data, as well as
a breadth of other freely available microstructure data that could ultimately support training
foundation market-making models. This study trades a single book at a time, and in
doing so lays the foundation for future work on collaborative multi-agent
learners that infer and exploit the constraints linking the many books of
such a venue \citep{saguillo2025arbitrage, kroer2016arbitrage}.

Given the structure of the problem, it is unsurprising that reinforcement
learning (RL) has been a paradigm shift for this literature. Cast as a
partially observable Markov decision process, single-agent RL market making
inherits the complexity of the underlying system while introducing its own
family of challenges: environment design, reward specification, and
often-times degenerate policies. The degeneracy that dominates this
environment is the collapse to no trading, in which the agent learns not to
quote, since inaction incurs a direct zero-variance return path that is
otherwise costly to attain through exploration
\citep{mohl2025jaxmarlhft, wang2023quote}. Two further difficulties
shape this study. Credit assignment is temporal: a quote posted now has
consequences that arrive only when the position it opened is closed, many
steps later, thus disentangling the temporal credit assignment problem is non-trivial. And the
environment itself is a modelling decision: simulators that model the market using Brownian mid-prices, fill probabilities, synthetic counterparties, tend to
understate adverse selection in ways that may result in an overestimation of performance, widening the Sim2Real gap, a widely acknowledged limitation
in the literature \citep{noble2026reality, lalor2024adverse}, while the venues we trade publish only book updates,
not the order-level data a faithful replay simulator needs.

More recent multi-agent work has shown evidence of adversarial training
improving the performance of market-making agents relative to training alone
or against static rule-based participants \citep{spooner2020robust,
gasperov2021signals}. In line with \citet{shoham2007if}, ``if multi-agent
learning is the answer, what is the question?'', our agenda here is
two-fold, both prescriptive and computational: we adopt multi-agent
reinforcement learning as an approach to market making in adversarial
settings, which in turn requires careful consideration of agent behaviour and
learning under complex, partially observable stochastic game (POSG) conditions \citep{becker2024posg}.

Building on the novel paradigm of deep reinforcement learning at scale
\citep{lu2022discovered}, we use JaxMARL-HFT
\citep{mohl2025jaxmarlhft}, a GPU-accelerated multi-agent environment that
replays recorded order flow through a full matching engine and cuts
end-to-end training time by up to 240 times against the reference
implementations they compare to, ABIDES-MARL \citep{cheridito2025abidesmarl}
and PyMarketSim \citep{mascioli2024pymarketsim}. The scope of their research was predominantly focused on framework design, while we focus on determining
improvements to their baseline market making performance in the adversarial setting. We address two
obstacles in particular, noted in their paper. The first is the no-trade collapse, which we treat
as a property of the environment design, in particular the reward and the action space, rather than of the learner.
The second is the entanglement of quoting and inventory management in one
policy, which we disentangle into a cooperative multi-agent desk: one agent
quotes continuously, the other manages the quoter's inventory position, and a reward split we control divides
the desk's profit between them. The aim of the study is to design an
environment and setting conducive to legitimate market-making learning
across thinly liquid and volatile markets.

We use two methods to evaluate our approach. Firstly, our experimental setup scores
every model on held-out (i.e. not in training) end-of-episode portfolio value. This is compared against
the baseline market maker configuration of JaxMARL-HFT, retrained on the same data. 
Our configurations are built to isolate the effect of each design choice and thereby gain evaluative insights; what we tried and did not pursue at scale is stated, with justifications, in \cref{sec:methodology}.

Our contributions are:
\begin{itemize}
\item \textbf{An enriched observation space}, giving the market maker the
  book and its recent flow rather than the spread and its own inventory
  (\cref{sec:obs}).
\item \textbf{A wider action space}, admitting skewed and one-sided quotes
  alongside the no-trade action (\cref{sec:actions}).
\item \textbf{Inventory control outside the reward}, through a cap and a
  loss stop rather than a penalty, so that the reward does not pay for
  inaction (\cref{sec:reward-enhancements}).
\item \textbf{A multi-agent decomposition of the market maker} into a quoter
  and a risk manager, trained as independent learners, giving inventory risk
  its own agent and objective (\cref{sec:actions-desk,sec:multiagent}).
\item \textbf{A reward design for the desk}: a general-sum split, a blend
  toward the desk total, and a markout term that charges the quoter for the
  mid's move after each fill, which together mitigate the credit-assignment
  problem the split creates (\cref{sec:reward-desk}).
\item \textbf{Order-level data for prediction markets}, inferred from
  Polymarket's L2 book updates and written in LOBSTER format
  \citep{huang2011lobster} (\cref{sec:data}).
\end{itemize}

The remainder of the paper is organised as follows. \Cref{sec:related}
reviews the stochastic-control, single-agent and multi-agent literature pertaining to market making.
\Cref{sec:background} defines our problem, including the environment, the state,
observation, action and reward spaces. It also includes our own refinements used to address challenges pervasive in the literature. \Cref{sec:methodology} presents the
learners, single-agent and multi-agent. \Cref{sec:setup} gives the
experimental setup alongside assumptions, simulator, data, models and settings
and \cref{sec:results} discusses the outcome of our experimentation. The \cref{sec:conclusion} conclusion sets out limitations and
future work.

\newsection{Related Work}\label{sec:related}

The central market maker problem arises from the endogenous and thereby
stochastic nature of market prices. As market makers aim to capture and
thereby profit from the bid-ask spread, there is no guarantee of both of
their quotes getting filled, which may result in the accumulation of unwanted
inventory \citep{fodra2012high}. With adverse price movements in the market
there is uncertainty regarding the future price of this held inventory 
(inventory risk \citep{ohara1986microeconomics}) thus requiring careful
management of filled positions. The issue is further exacerbated by the
adverse selection problem observed by \citet{glosten1985bid}: informed traders
exploit private information unknown to the market maker, and this information
asymmetry produces adverse fills as the price rallies in a given direction.

\subsection{Stochastic control problem}

Modelling the theoretical problem of market making as a stochastic dynamic
program, \citet{ho1981optimal} began the approach of deriving optimal bid-ask
quotes for a monopolistic dealer in a single stock that maximise an objective
function: the utility of the market maker's expected terminal wealth. A
significant refinement of this approach was offered by
\citet{avellaneda2008high}, who combined this utility framework with the microstructure of limit order books. In particular, they
leveraged notable results from the econophysics literature pertaining to the
size distribution of market orders \citep{maslov2001price, weber2005order,
gabaix2006institutional} and their temporary price impact
\citep{bouchaud2002statistical, weber2005order} in order to model the arrival
of buy and sell orders as Poisson processes tied to the market maker's
quotes, founded on \citet{garman1976market}. They first compute the agent's
closed-form reservation price, a mid-price adjusted with respect to held
inventory to reflect indifference, and then derive closed-form expressions
for the optimal bid and ask quotes as distances from that reservation price.

Crucially, this framework relies on parametric, memoryless Poisson rate
processes, creating a distinct Sim2Real gap. While mathematically
tractable, such models inevitably abstract away real-world
microstructural features such as queue dynamics, latency and volatility
clustering, and this reality gap in limit order book simulation leads to
distorted execution costs and profit profiles that might trivialize the
performance of the market maker in real-world scenarios. \citep{noble2026reality}.
\citet{lalor2024adverse} show the sharper consequence: simulating prices and
order flow independently of one another omits endogenous adverse fills,
thus inflating apparent profitability of such methods, and performing in an incommensurate manner in live markets.

To bridge this gap while preserving mathematical tractability, subsequent
literature expounded on the underlying asset dynamics in order to capture more complex,
observable market states. \citet{gueant2013dealing} solved the market making
problem under inventory constraints across arbitrary time horizons using a
stochastic control change of variables. \citet{fodra2012high} further extended
these frameworks to a broader class of mid-price models, including
arithmetic Brownian motion with drift and mean-reverting Ornstein-Uhlenbeck
processes, under both exponential and linear utility functions, while
introducing a terminal inventory-risk penalty. These solutions remain bound to
the model that makes them solvable, given that every quote is optimal with respect to the
assumed dynamics, not the observed ones. To bypass this bottleneck of
restrictive closed-form solutions, the quantitative literature has been
converging toward reinforcement learning \citep{bai2024review}.

\subsection{Reinforcement learning}

\citet{chan2001electronic} were the first to formalise market making as a
reinforcement learning problem. Their agent's state includes both its own
information such as inventory, as well as market characteristics including the imbalance
of buy and sell orders, the bid-ask spread and price changes between trades.
Its discretized action space allows the agent to set quote prices and sizes, while its
reward mixes profit with an inventory penalty. They populate their simulated
market with informed and uninformed traders whose orderflow responds to
the maker's quotes, so the elements of the problem we study, namely: partial
information, informed flow, and a reward balancing profit against inventory, are all present in that formulation. This laid
the foundation for further exploration of RL for MM.

Subsequent single-agent literature explores how risk management and different learners within the broad framework
established by \citet{chan2001electronic}. On the value-based side, \citet{mani2019applications}
replace a SARSA learner with double SARSA as well as a customised
temporal-difference update for the Q-function, reducing the variance of the reward and reaching higher profit at lower inventory than the policy of
\citet{chan2001electronic}. \citet{spooner2018market} leverage tile-coding to discretize the action space and reshape the reward itself with an
asymmetrically dampened profit term paying the agent for providing
liquidity rather than for speculating with inventory. This reward is inherited by our baseline configuration (\cref{sec:reward-basic}).
\citet{selser2021optimal} instead explore a penalization term for the estimated variance of the agent's wealth, demonstrating that
 a deep Q-network under this reward outperforms the
closed-form benchmark of \citet{avellaneda2008high}. Moving from value-based
learners to policy gradients, \citet{gueant2019deep} use an actor-critic
method for multi-asset market making and scale it to a hundred corporate
bonds, a dimensionality the partial-differential-equation methods cannot
reach (this lays the foundation for future work pertaining to scaled multi-agent systems across combinations of loosely correlated prediction market assets). 
We set out our own reasons for preferring the on-policy policy-gradient family
over value-based learning in \cref{sec:pg}.

These methods differ as much in their environments as in their learners. Most
train inside model-driven simulators, with a Brownian or mean-reverting
mid-price with fill probabilities supplied by a model, similar to what was discussed in the previous section, where fills are cheap and adverse selection
is a consequence of the rigour of the modelling. On the other hand, on real, replayed order flow fills are disproportionately informed, and a maker trained
there often learns to stop quoting altogether, the collapse JaxMARL-HFT
documents and that we address in \cref{sec:notrade} \citep{mohl2025jaxmarlhft}.
\citet{wang2023quote} reach a consistent conclusion from the model-driven
side: relaxing the standing assumption that a maker quotes both sides
continuously and instead allowing it to abstain, or to quote one side only, improves
returns and risk-adjusted returns. This justifies our explicit no-trade and one-sided actions in our action space for the market maker.
(\cref{sec:actions}).

\subsection{Multi-agent reinforcement learning}
Reinforcement-learning for market making has been extended to several agents in three broad directions \citep{bai2024review}. The first is adversarial:
\citet{spooner2020robust} train the maker against an adversary controlling
the parameterization of the market model itself, allowing the learned policy to become
robust to model misspecification; \citet{gasperov2021signals} let the
adversary displace the maker's quotes directly, reducing inventory risk and
improving profit relative to training without one. \citet{spooner2020robust}
deliberately choose market dynamics modelled by the optimal-control approach,
allowing them to conduct clean single-stage analysis and avoids the
biases of historical data. Our choice is precisely the opposite, we accept 
the analytical inferiority related to recorded data and instead focus on improving the performance
against real flow, given our focus is on real performance.
The second direction is competitive:
\citet{ganesh2019reinforcement} place the learner in a dealer market against
competing market makers, showing that it learns their pricing distributions
from only their own observations, thereby pricing near-optimally against competitors whose
parameters it never sees. The third decomposes the market-making decision:
\citet{patel2018optimizing} splits this hierarchically between a macro-agent
choosing direction and a micro-agent placing the orders in the book. Our desk
(\cref{sec:reward-desk}) is also a decomposition of one seat into two learners, a
quoter and a risk manager, but the two act simultaneously as peers on real
flow, under a reward split we control, instead of in a hierarchy.

The environments these systems train in divide the same way as the
single-agent ones, and the division decides what a result means. Agent-based
discrete-event simulators, most prominently ABIDES \citep{byrd2020abides} and
its reinforcement-learning interface \citep{amrouni2021abidesgym}, populate a
matching engine with modelled background traders: the market reacts to the
agent, which cannot be said for historical replay, but on the other hand every counterparty is synthetic, so
what the agent learns is bounded by the quality of the background agent's imitation of real flow.
Optimal-control markets model the dynamics in the manner previously described.
JAX-LOB \citep{frey2023jaxlob} takes a different path: it replays recorded order
flow, order message by order message, through a full price-time-priority matching
engine, and JaxMARL-HFT \citep{mohl2025jaxmarlhft} extends this approach to several
learning agents, building on JaxMARL \citep{rutherford2024jaxmarl}. We thus chose to build
on JaxMARL-HFT for two reasons. The first is realism: fills come from
recorded orders resting in a real orderbook rather than from a fill-probability
model or synthetic agent-based counterparties, so the adverse selection faced by our agents is reflective of the markets itself. 
The second is scale: ascribing to the hyper-scale paradigm recently unlocked by this line of work \citep{lu2022discovered,
rutherford2024jaxmarl}, the engine and learner compile as one
JAX program, so thousands of books can step in parallel on a single GPU. This is what made our experiment matrix and broader experimentation possible.

Within this environment, our taken direction is hitherto untried: a cooperative
decomposition of the market-making seat itself. In line with the new unlock of reinforcement
learning at hyper scale, we aim to improve upon the market maker proposed in
JaxMARL-HFT, offering a refinement within its simulation environment that
might further the research herein.

\newsection{Preliminaries}\label{sec:background}

This section describes the pre-requisite foundations that the method is predicated on: how a limit order book works,
what the offered prediction market assets/contracts are, and the four factors that define our learning problem: the state, the observations, the actions and the reward.
\Cref{sec:methodology} builds directly on the formalism introduced here.

\subsection{Limit order books and binary contracts}\label{sec:lob}

An exchange matches buyers and sellers of a given asset through a limit order book. A limit
order names a side (bid/ask), a price and a size, and rests in the book until it gets filled (i.e. trades)
or is cancelled. A market order trades immediately against the best resting
orders on the opposite side, identical to an immediate-or-cancel (IOC) order, with the added feature that IOC's will discard any quotes not traded immediately. 
Limit orders resting at the same price get filled in the order that they arrived, it is in this sense that the book runs under price-time
priority. Polymarket uses a central limit order book (CLOB) with a matching engine
that uses price-time priority, thus aligning with our own matching engine. At time $t$ the highest resting buy price is the best bid $b_t$ and
the lowest resting sell price is the best ask $a_t$. Their midpoint
$m_t = (a_t + b_t)/2$ is referred to as the mid price and $a_t - b_t$ is referred to as the spread. Prices move at minimum increments called ticks (these are stated for each market on Polymarket).

A market maker quotes a bid and an ask price at the same time. If both fill it earns
the distance between them (spread), if only one fills it holds a position that moves with the value of the mid (mark-to-market). 
Its profit therefore has two parts: the spread captured on
completed round trips, and the gain or loss on inventory carried between fills.
The second part is the more challenging to manage. As previously established, a fill occurs because another participant chose to trade at the price quoted by the market maker, 
and the participants timing their trades prior to mid price moves based on private information are those a market maker would like to avoid given the potential for adverse selection\citep{glosten1985bid}. 
The quality of a fill, and of the market maker's quoting, therefore is predominantly revealed by the subsequent movement of the mid-price.

The contracts offered on Polymarket are binary: each pays one unit if a named event occurs by the resolution date
and otherwise pays nothing. It is similar in this sense to a Cash-or-Nothing digital option, except that the trigger condition is not necessarily a function of an underlying asset price. Its price sits in the unit interval with the minimum tick
$\delta$ as the discrete increment,
\begin{equation}
  p \in \{0,\, \delta,\, 2\delta,\, \ldots,\, 1\} \subset [0,1],
\end{equation}
and reads directly as the market's implied probability of that event. It resolves at
a known deadline to zero or one. One consequence matters: time to
resolution counts as much as price, given time-decay characterises these markets.

\subsection{The decision model}\label{sec:decision-model}

Market making is a sequential decision problem under uncertainty. At each step
$t$ the environment holds a state $s_t$, the agent receives the observation
$o_t$, choosing an action $a_t$. The environment accordingly returns a reward $r_t$ and
then updates its state to $s_{t+1}$.

In our context, the agent does not see the state. Since the state contains every resting order, every
market participants' intentions, and the private information that the other market participants act on,
while the agent sees the book and other quantities computed from it. Each learner is trained under the single-agent view of the market, thus it is in this sense that we model our learning as a partially observable Markov decision process (POMDP) \citep{kaelbling1998planning}.
The agent's observations are obtained by treating the other participants as part of the environment, in that the observations only hold a partial view of the entire
market state. Thus, the agent cannot act on exclusively $o_t$. Instead, rather than using the
full history $h_t = (o_0, \ldots, o_t)$, our policy conditions on a recurrent
summary of it, which consists of a learned approximation of the distribution over states implied by the history. Given the other learners are simultaneously adapting
to each other's policies, the environment faced by the agents is not stationary.

The partial observability here is inherent to the endogenous price formation in markets. The
information asymmetry that generates the order flow is itself the unobserved
part of the state, and inferring how informed the flow arriving now is precisely
the substance of the market maker's problem. This holds against the replayed
participants but not within our desk configuration since the two agents of \cref{sec:obs-desk} are
given each other's positions by construction.

Forcing the agent to close its position at every step would make each step's profit self-contained and restore a
memoryless problem, but it would also inevitably restrict the set of policies that the agent
can represent, and the best policy of the unrestricted problem, which holds a
position while it works, may not lie inside the restricted set. We therefore let
inventory carry across steps, thus requiring a more careful attention to the temporal credit assignment challenge: a quote placed
now has consequences that appear only when the resulting position closes,
possibly many steps later.

Given our context, we clear the recurrent summary at every episode boundary. In particular, this is done given that
we have modelled each episode as one market window, and the environment subsequently draws the next window at random instead 
of continuing on the same market, so clearing ensures that the recurrent state does not carry over a history from an unrelated market.

\subsection{The simulator}\label{sec:sim}

Everything in this work runs in the aforementioned simulation environment, on two frameworks. The matching engine is JAX-LOB \citep{frey2023jaxlob}: a limit order
book written in JAX.  JaxMARL-HFT \citep{mohl2025jaxmarlhft} inherits the matching engine to supply
multi-agent reinforcement-learning environment. Several agents learn on one
book simultaneously. Each agent's action is converted into an order message, those
messages join the recorded fixed stream of messages for a given timestep, and the matching engine
processes this combined stream; \cref{fig:desk-step} illustrates one such step in our simulator. From
the subsequent state, the environment computes agent's reward and subsequent
observations.

\begin{figure}[t]
\centering
\begin{tikzpicture}[
  font=\sffamily\footnotesize,
  agent/.style  ={draw=ag, fill=ag!14, rounded corners=2pt, minimum width=25mm,
                  minimum height=7.5mm, align=center, font=\sffamily\footnotesize},
  seat/.style   ={draw=ag, densely dashed, rounded corners=3pt,
                  inner xsep=2.6mm, inner ysep=3mm},
  envbox/.style ={draw=tp, fill=tp!10, rounded corners=3pt,
                  inner xsep=3.5mm, inner ysep=3.5mm},
  stage/.style  ={draw=black!45, fill=black!3, rounded corners=2pt,
                  minimum height=7mm, align=center, font=\scriptsize},
  state/.style  ={draw=black!50, fill=st!45, rounded corners=2pt,
                  minimum height=7mm, align=center, font=\scriptsize},
  engine/.style ={draw=en, fill=en!22, rounded corners=2pt,
                  minimum height=9mm, align=center, font=\scriptsize},
  seg/.style    ={draw=ag, fill=ag!18, minimum height=5.5mm, align=center,
                  font=\scriptsize, inner xsep=1pt},
  tapeseg/.style={draw=black!55, fill=white, minimum height=5.5mm, align=center,
                  font=\scriptsize, inner xsep=1pt},
  flow/.style   ={-{Latex[length=1.8mm]}, draw=black!65},
  agflow/.style ={-{Latex[length=1.8mm]}, draw=ag, thick},
  tpflow/.style ={-{Latex[length=1.8mm]}, draw=tp},
  lbl/.style    ={font=\scriptsize, text=black!60, align=center},
  key/.style    ={font=\scriptsize, text=black!55, align=left}
]

\node[agent] (quoter) at (1.45,0.55)  {quoter};
\node[agent] (risk)   at (1.45,-0.45) {risk manager};
\node[seat, fit=(quoter)(risk), label={[font=\sffamily\scriptsize\bfseries, text=ag]
      above:market maker}] (seat) {};
\node[agent, draw=ag!55, fill=ag!4, text=black!60] (exe) at (1.45,-1.95)
      {execution agent};

\node[state, text width=80mm] (st) at (9.3,1.25)
   {\textbf{state $s_t$}: the order book, every resting bid and ask,
    and each agent's inventory and cash};

\node[lbl, anchor=south] at (9.3,0.12) {one stream of messages:};
\node[seg, anchor=west, minimum width=36mm] (sA) at (5.4,-0.35)
     {the agents' messages};
\node[tapeseg, right=1.4mm of sA, minimum width=44mm] (sT)
     {this step's recorded market messages};

\node[engine, text width=78mm] (eng) at (9.3,-1.75)
   {\textbf{JAX-LOB matching engine}: applies each message to the orderbook, matching against what rests there, and records the subsequent trades};
\node[state, text width=86mm] (st1) at (9.3,-3.1)
   {\textbf{state $s_{t+1}$}: carries both sides of the book, this step's trades, the
    time to resolution, and each agent's inventory, cash and open positions};

\draw[flow] (st.south) -- (9.3,0.55);
\draw[flow] (9.3,-0.68) -- (eng);
\draw[flow] (eng) -- (st1);

\begin{scope}[on background layer]
\node[envbox, fit=(st)(sA)(sT)(eng)(st1),
      label={[font=\sffamily\scriptsize\bfseries, text=black!60]above:environment}]
      (env) {};
\end{scope}

\draw[agflow] (seat.east) -- node[lbl, above, pos=0.5] {actions $a_t$}
              (seat.east -| env.west);
\draw[agflow, draw=ag!55] (exe.east) -- (exe.east -| env.west);
\draw[flow] (env.south) -- (9.3,-4.35) -- (-0.8,-4.35)
            -- (-0.8,0.05) -- (seat.west);
\node[lbl, anchor=south, fill=white, inner sep=1.5pt] at (4.6,-4.29)
   {observation $o_{t+1}$ and reward $r_t$};

\node[key, anchor=north west, text width=145mm] at (-0.8,-4.75)
   {Each agent's message is comprised with a withdrawal of what it no longer wants resting (i.e. quotes to cancel) and posts
    its preferred quotes. A quote repeated at the same price and size of a past quote is left in
    place to retain its queue priority. The agent messages are shuffled between them before entering the matching engine, so no
    agent has systematic priority over another (i.e. no model of latency advantage).};

\end{tikzpicture}
\caption[One step of the environment]{One step. All agents get their observations of the state and act on it simultaneously, receiving their reward from
their actions. The desk is interchangeable with the single-agent, where the single-agent would perform both jobs. A step is a fixed number of messages rather
than a fixed duration, meaning its length in market time varies by how busy the
market is.}
\label{fig:desk-step}
\end{figure}

\subsection{State space}\label{sec:state}

The state collects two parts: a shared description of the market, and a private
record for each agent.

The shared part is the state described in \citet{frey2023jaxlob}:
both sides of the resting orderbook, each as an array of orders carrying a price,
size, timestamp and an identifier for the trader who placed it. The trades executed
during the step, the episode's start time and the current clock, a counter that
issues identifiers to new orders, and finally the index of the market being traded
together with the step counter. JaxMARL-HFT extend this with three quantities a
market maker needs: the path of the best bid and
ask after every message of the step, the mid price, and the amount of time that has been covered by the steps. Our own addition is smaller and serves the desk: four fields
through which an agent type can read the other agent's position, which is used by 
\cref{sec:obs-desk}.

The private part, per agent, holds the signed inventory, the cash balance, the
running profit, the volume-weighted price at which the open position was
entered, the distances of the agent's own resting quotes from the mid, and two
short histories used to build features, recent mid prices, and a per-step bar
of open, high, low, close, volume and trade imbalance. The desk adds a table of
the open positions: each row carrying a signed size, the price it was opened at,
and its age in steps.

Everything that changes the book arrives as a message,
\begin{equation}
  \mathbf{m} = \big[\, T,\; S,\; Q,\; P,\; \mathit{OID},\; \mathit{TID},\;
                  T_{s},\; T_{ns} \,\big],
  \label{eq:message}
\end{equation}
giving the type $T$ i.e. a limit order, a cancellation, a deletion or a market
order, the side $S$, the size $Q$, the price $P$, an identifier for the order and the corresponding trader, and the time in seconds and nanoseconds. The trader identifier is
what lets each fill be matched to the corresponding agent. See \citet{frey2023jaxlob} for the full
specification.

The agent observes none of this directly, but rather observes a vector built from these quantities.

\subsection{Observation space}\label{sec:obs}

The market maker of JaxMARL-HFT observes the spread and its own
inventory. Given that this limited observation gives no substantial information to the agent as to the nature of the book and possible movements of the mid-price, the policy can only condition on 
its own exposure but not the dynamic market its quoting into, which we note as a significant limitation.
We accordingly chose a richer observation space in order to remove that
limit, it is comprised of three blocks.

\paragraph{The book.} For each of the five levels $i = 1, \ldots, 5$ on each
side of the book, the observation carries the distance from mid, in ticks,
and the size on each level.
\begin{equation}
  o^{a}_{i,t} = \frac{p^{a}_{i,t} - m_t}{\Delta},
  \qquad
  o^{b}_{i,t} = \frac{m_t - p^{b}_{i,t}}{\Delta},
  \qquad
  v^{a}_{i,t}, \; v^{b}_{i,t},
  \label{eq:obs-book}
\end{equation}
where $v^{a}_{i,t}$ and $v^{b}_{i,t}$ are the total resting size at that given price.
A level that does not exist is reported as zero.

\paragraph{The flow.} Order imbalance indicates which side has a larger density of resting orders, thereby signalling possible movements in mid price
\begin{equation}
  \iota_t = \frac{\sum_i v^{b}_{i,t} - \sum_i v^{a}_{i,t}}
                 {\sum_i v^{b}_{i,t} + \sum_i v^{a}_{i,t}}.
  \label{eq:obs-flow}
\end{equation}
Two further terms are used to describe recent prices: a relative-strength measure over the last
mid prices, as well as the standard deviation of their step-to-step changes in ticks.
A summary of the step is added to these: the range, return and
closing price of the step's high-low bar, its share of traded volume
that did not get filled, the drift across the window, and the average of the
signed shares over the step.

\paragraph{The agent's own situation.} Inventory $q_t$, the spread, the step
counter, and, given the theta decay inherent to this market, the fraction of the
episode remaining. The open position is marked against the mid.

This tells the agent whether its current holdings are winning or losing (not covered by the raw inventory).

We accept the much wider observation space given the structural blindness of the alternative (spread + inventory). In addition, this exact space was the conclusion of 
preliminary experimentation.

\subsubsection{Observations in the desk}\label{sec:obs-desk}
The desk is built in order to mirror the separation of work in real market making desks. One quoting agent and another managing the underlying risk.
We refer to them as the quoter and the risk manager.

Two agents sharing one book would have to coordinate, and a possible approach to this would be
communication actions, or shared parameters or messages. We do not take that approach given concerns about multi-agent credit assignment 
and instead \emph{augment} each agent's observations with a block describing their partner's actions. Coordination is therefore
expected to be inferred from what the partner's actions together with the reward function design. This implicit communication
required the following choices in the observation space.

We denote $o^{\text{mkt}}_t$ as the observations in \cref{sec:obs}: the market,
plus the agent's own inventory and position mark, which both desk agents
receive. The desk's two observations are then
\begin{align}
  o^{\text{quoter}}_t &= \big[\; o^{\text{mkt}}_t,\;\;
                                 o^{\text{own}}_t,\;\;
                                 o^{\text{closed}}_t \;\big],
    \label{eq:obs-quoter}\\[2pt]
  o^{\text{risk}}_t   &= \big[\; o^{\text{mkt}}_t,\;\;
                                 o^{\text{net}}_t,\;\;
                                 o^{\text{own}}_t(\text{partner}),\;\;
                                 o^{\text{lots}}_t \;\big],
    \label{eq:obs-risk}
\end{align}
Where $o^{\text{own}} \in \mathbb{R}^{4}$ are the agents positions split into long and short
units with their entry offsets, and for the quoter $o^{\text{closed}} \in \mathbb{R}^{4}$ is the
units and profit the risk manager closed in the previous step (an implicit form of communication/feedback).
For the risk manager, $o^{\text{net}} \in \mathbb{R}^{2}$ is the quoter's inventory and the desk's
net position (i.e. the combined position), and $o^{\text{lots}} \in \mathbb{R}^{8 \times 4}$ is the quoter's open-position table. 
The environment records at most eight open positions, each row of the table holds one of these positions. Rows without a position are zeroed such that the volume is zero so the agent cannot conflate it for a real open position.
Each row carries four numbers: the position's size $s_r$, its entry offset from the mid $(p_r - m_t)/\Delta$, the number of steps since it was opened, and its mark against the mid,
\begin{equation}
  \pi_{r,t} = \frac{s_r\,(m_t - p_r)}{\Delta},
  \label{eq:obs-row}
\end{equation}
with the positions ordered worst to best such that $r = 1$ is always the one
losing most. The risk manager's view is thus more tailored to inventory and risk management in a manner that 
the quoter does not have. Together with the action space and the reward design, this allows us to break down the market making problem
into a tractably richer problem with more narrow focuses on each underlying challenge to market making.

\subsection{Action space}\label{sec:actions}

The design of the action space is non-trivial as it fixes the range of possible policies available to the agent. 
A discretized space of possible quote placements, for instance, can only express
policies built from those placements. This implies that learning can then converge only to
the best of that constricted action space. In contrast, a continuous space of offsets (against the mid)
allows the exploration of a broader set of candidate policies: one that may contain the game's
global optimum (subject to the constraints of the problem and the observations). As such,
a richer action space is more likely to contain a globally superior policy evaluated on portfolio value, but the 
cost incurred is learning, in particular that a larger space needs more interaction data and makes approximation much more challenging.

We thus settled on two choices inside this trade-off. Firstly, we kept the action space discrete. 
The reasoning for this choice was two-fold. Firstly, the large size of our observation space 
paired with a continuous action space would have likely made the learning intractable, which corroborates
the pervasive use of discretization in the literature. Secondly, with a wider and sufficiently granular discretization (namely tick-level), which is afforded to us
by the abundance of Polymarket data and the JAX accelerated framework, we can attain a comparatively rich action space without requiring full continuity. 

\subsubsection{The quote table}\label{sec:quote-table}
At each step the agent selects one row $k$, carrying a bid
offset, an ask offset and a size on each side, from the following table:
$(\delta^{b}_k, \delta^{a}_k, q^{b}_k, q^{a}_k)$, and the offsets are applied to
the best bid/best ask in ticks,
\begin{equation}
  p^{b}_k = b_t - \delta^{b}_k \Delta,
  \qquad
  p^{a}_k = a_t + \delta^{a}_k \Delta,
  \label{eq:quote-table}
\end{equation}
at sizes $q^{b}_k$ and $q^{a}_k$ denominated in units of the contract. The best bid
and ask in \cref{eq:quote-table} are recomputed with the agent's own resting
orders removed, ensuring it does not quote relative to itself. A row with a size of
zero on a side sends no message on that side at all, rather than an order of
zero size. That is the mechanism behind both the one-sided rows and the row
that posts nothing.

The table given to the agent dictates its range of possible actions.
JaxMARL-HFT's action table (that performed the best) has nine rows, every one two-sided, reaching at most
five units from the best bid and best ask. We propose a
wider table of seventeen rows, written as offset pairs
$(\delta^{b}, \delta^{a})$ with both sides at unit size unless a size is
shown:
\begin{itemize}
  \item \emph{symmetric} (seven):
        $(0,0),\ (1,1),\ (2,2),\ (4,4),\ (6,6),\ (9,9),\ (13,13)$;
  \item \emph{skewed} (six), pairing a near side with a far one:
        $(0,2),\ (2,0),\ (1,4),\ (4,1),\ (2,6),\ (6,2)$;
  \item \emph{one-sided} (four), in full
        $(\delta^{b}, \delta^{a}, q^{b}, q^{a})$ form:
        $(0,0,1,0),\ (0,0,0,1),\ (4,0,1,0),\ (0,4,0,1)$ --- the zero size
        is the instruction not to quote that side.
\end{itemize}
While a market maker is expected to show both sides, there are states in which
quoting only one is a better and legal (with respect to exchange market making regulations) action: for instance a maker that is already long as the price
falls has no reason to keep bidding. The one-sided rows allow the
policy to express such a strategy. The two differences that matter (relative to JaxMARL-HFT) are the reach,
thirteen ticks against five, and the one-sided rows. Together with the row posting no trade appended, the head is ten actions wide for the baseline and
eighteen for ours.

\subsubsection{The no-trade action}\label{sec:notrade}
In JaxMARL-HFT, the trained market maker selects the no-trade action between 61.1\% and
97.7\% of the time across three seeds \citep{mohl2025jaxmarlhft}. We refer to this as the no-trade collapse.
We decide to keep this action row for three reasons. Firstly, standing aside is part of the market making role and should thus be reflected
in the action space of our agent. Even in exchanges
that have formally imposed market making obligations, this is legally permissible: the London Stock Exchange requires registered market makers to quote for at
least 90\% of continuous trading during the mandatory period, and Deutsche
B\"orse's cash market requires quotes during 50\% of daily trading hours on a
monthly average, with the requirement lifted in defined exceptional
conditions \citep{wang2023quote}. Our exchange has no regulation for market makers at all, so our context is especially validated.

Secondly, removing this action does not change the underlying behavior leading to the no-trade collapse. 
If inaction is what the reward rewards, the policy will simply search for inaction out of whatever the action space
provides. We deleted the row from the baseline and it did exactly that, moving
most of its action mass onto the widest symmetric rows and quoting several
ticks outside a book two ticks wide, where nothing fills --- an observation
from our preliminary experimentation.

Thirdly, the reward design is particularly responsible for making inaction a desirable policy 
JaxMARL-HFT attribute part of the collapse in their own agent to the inventory penalty, which charges the agent
for the size of the position it holds. That charge is smallest when the
position is zero and thus causes a term intended
to control risk to instead result in a reward for inaction \citep{mohl2025jaxmarlhft}.
The no-trade collapse is therefore a reward problem that we aim to address in \cref{sec:reward}.

\subsubsection{The desk action space}\label{sec:actions-desk}
A real trading desk typically comprises a quoting agent and a risk-management agent who operate on the same book. 
We give the two agents action sets that reflect this intuitive division of labour. The motivation for this separation has been aforementioned.

The quoter uses the table of \cref{sec:quote-table}, while the risk manager never quotes at all. Instead, its action space is defined by managing
the book of the quoter.

\begin{equation}
\begin{aligned}
  \mathcal{A}^{\text{risk}} =\;
  &\underbrace{\{\,\text{nothing}\,\}}_{1}
   \;\cup\;
   \underbrace{\{\text{a quarter},\ \text{a half},\ \text{all}\}
     \times
     \{\text{losing},\ \text{winning},\ \text{long},\ \text{short}\}}_{12} \\[2pt]
  &\;\cup\;
   \underbrace{\{\,\text{close position } 1, \ldots, 8\,\}}_{8}.
\end{aligned}
  \label{eq:rm-actions}
\end{equation}

Closing a share of one of the four groups (losing, winning, long, short) is
tantamount to liquidation in that the desk sells part of what it holds,
(or buys it back if the position is short) immediately. The action
becomes an immediate-or-cancel (IOC) order on the opposite side of the position,
priced at the best bid or best ask: it trades at once against what is
resting there, and any part that cannot trade is discarded rather than left
in the book. That order is stamped with the quoter's identifier instead of
the risk manager's, so it reduces the quoter's position directly. The eight
named positions are the rows of the quoter's book ordered worst to best, the same ordering described in \cref{eq:obs-row}.

This leads to two multi-agent credit-assignment problems. Firstly, the cost of crossing the spread to liquidate the position is charged to the quoter,
which did not choose to cross. Secondly, the implicit consequences (with regard to unrealised PnL) of the quoter's actions are reflected in the reward of the risk manager. It is in this sense that the reward design is non-trivial
(\cref{sec:reward-desk}), as it is there that we disentangle these opposing forces by allowing the agents to implicitly recognize the consequences of their own actions.

\subsection{Reward function}\label{sec:reward}

\subsubsection{The basic reward}\label{sec:reward-basic}

The objective of a market maker ought to be profit, so the natural reward is the profit and
loss its trades produce. This is the form adopted by \citet{spooner2018market}, and consequently by JaxMARL-HFT \citep{mohl2025jaxmarlhft}. 
It sums the realised edge on fills (relative to the mid price) as well as the unrealized marked-to-market inventory held.

$\mathrm{Buy}_t$ and $\mathrm{Sell}_t$ denote the executed volume/fills against the
agent's bid and ask since the previous step, at prices $p^{b}_t$ and $p^{a}_t$,
and $q_t$ for the held inventory after the step. The spread captured on fills,
measured against the mid, is
\begin{equation}
  \psi^{b}_t = \mathrm{Buy}_t \cdot \big(m_t - p^{b}_t\big),
  \qquad
  \psi^{a}_t = \mathrm{Sell}_t \cdot \big(p^{a}_t - m_t\big).
  \label{eq:fill-edge}
\end{equation}
Each term is the distance between the price the agent paid or received and the
mid at the time, so it is positive whenever the agent bought below the mid or
sold above it. The inventory the agent already held is revalued as the mid
moves, and the reward is the sum of the two,
\begin{equation}
  \Phi_t = q_{t-1}\big(m_t - m_{t-1}\big),
  \qquad
  r_t = \psi^{b}_t + \psi^{a}_t + \Phi_t.
  \label{eq:base-reward}
\end{equation}

This is aligned with the quantity the agent is judged on: its portfolio
value $V_t = c_t + q_t m_t$ is its cash plus its inventory marked at the mid. Summed over an episode, \cref{eq:base-reward}
amounts to $V_T - V_0$, meaning that the more granular per-step reward and the end-of-episode evaluation measure agree.
Our implementation marks fills against the step's average mid, which is the more appropriate price for a fill that
may have happened at any point throughout the step.

\subsubsection{Where the basic reward falls short}\label{sec:reward-problems}

The form is simple and matches the real-world objective, but contains three properties that encourage the agent to converge
to policies we do not want.

\begin{itemize}
  \item \textbf{It is dominated by price noise.} The revaluation term follows
        the step-to-step movement of the mid and thus the variance of this unrealised PnL term grows with the position. This causes the reward to be dominated by short term price fluctuations that dampen the 
        valid signal telling the agent whether it quoted appropriately or not.
  \item \textbf{It does not charge for inventory magnitude.} Nothing in \cref{eq:base-reward}
        costs the agent anything for holding a large position beyond the price moving against it, so inventory can accumulate until the mid price move is significant, causing greater reward volatility.
  \item \textbf{It mishandles temporal credit assignment.} The spread capture is rewarded in \cref{eq:fill-edge}, yet the identical action's temporal consequences a few steps out are inside $\Phi_t$. With incommensurate weights, the agent may become more averse to risks taken in the present given it is not sufficiently rewarded for the consequences of the action a few steps later. Thus, it naturally may collapse into "No Trade".
\end{itemize}

The standard remedy for the second problem, an inventory penalty, further exacerbates the other two stated problems. By charging the agent for held inventory,
the agent during learning may opt to drive that charge predictably to zero as a possible policy. The mid moves on almost every step, so the
revaluation term is noisy provided the agent holds any inventory. Therefore a policy that quotes anything at all would receive both that noisy mark-to-market and be further
exacerbated by the penalty: encouraging the policy to stand aside. 

\subsubsection{Enhancements}\label{sec:reward-enhancements}

The literature's standard responses to these problems act through the reward
itself. \citet{spooner2018market} damp the inventory term asymmetrically:
with $\gamma \in (0,1]$ and $\eta \in [0,1)$,
\begin{equation}
  r_t = \psi^{b}_t + \psi^{a}_t
        + \gamma\Big(\Phi_t - \max\big(0,\ \eta \Phi_t\big)\Big),
  \label{eq:asym-damped}
\end{equation}
so that a favourable revaluation is credited at less than an adverse one is
charged, and holding inventory stops being a source of profit, thus encouraging direct profits from spread capture. The quadratic
inventory penalty $-\nu q_t^{2}$ charges inventory directly. Both,
however, inherit the difficulty the previous subsection describes given that any charge
routed through the reward for holding inventory is smallest at zero
inventory thereby leading to the degenerate policy.

We therefore control inventory outside the reward instead:

A \textbf{hard inventory cap} prevents the position from exceeding a fixed number of units, thus limiting 
inventory without implicitly discouraging trading in the reward function. This ensures a more stable reward variance and discourages the "No Trade" collapse, at the expense
of a more constrained set of feasible actions.

Secondly, we have a \textbf{loss stop with an annealed threshold}. When an
episode's accumulated loss (measured in PnL terms) crosses a given threshold, the agent stops quoting new positions and instead
is restrained to closing its current positions. The
threshold starts loose and tightens over training in order to ensure that the policy can learn through trading early on
and progressively learn a more disciplined method of inventory management later.

JaxMARL-HFT propose one further remedy for the no-trade collapse, namely having the reward be scaled as a function of the volume traded by the agent, 
thereby reducing the incentive for inaction. Having tested this approach to no avail, we opt for above two solutions.

Despite these solutions, the third problem of temporal credit assignment remains, which in turn motivates our multi-agent formulation.

\subsubsection{The reward in a multi-agent desk}\label{sec:reward-desk}
A single agent holding one reward is required to trade off the quality of their quoting against the feedback
on their inventory carry (often referred to as unrealised PnL). Given the two are measured in the same reward, the gradient update cannot separate the two practically different objectives/terms.
Thus, splitting the desk gives each
agent a smaller problem with a cleaner objective, making the problem both more tractable
and closer to the intuitive separation of the roles.

Rewarding both agents in a common-sum manner following the same single-agent reward would detract from the diverging policies and feedback we wish the respective agents to leverage. Yet rewarding each of them
exclusively for spread capture and unrealized PnL would equally create moral hazard problems. 
namely, for instance a quoter rewarded for spread capture reinforces quoting without accounting for the quality of the position, with the cost of the carry accruing exclusively to the risk manager.

\paragraph{The split.} The quoter is rewarded the spread capture on its filled
quotes. The risk manager is paid the unrealised PnL of the positions those
fills open (i.e. the inventory marked-to-market), and is penalized with the cost of its own closing orders (to ensure it does not collapse to a degenerate policy of always liquidating all inventory). Both rewards are measured in ticks. Writing
$\psi^{b}_t + \psi^{a}_t$ for the spread captured, $\Phi_t$ for the
unrealised PnL and $c_t \ge 0$ for the cost of the closes:
\begin{equation}
  \tilde r^{Q}_t = \psi^{b}_t + \psi^{a}_t,
  \qquad
  \tilde r^{R}_t = \Phi_t - c_t.
  \label{eq:desk-split}
\end{equation}
The two rewards sum to the desk's change in portfolio value.

\paragraph{The blend.} \Cref{eq:desk-split} leaves one incentive untreated: the quoter is paid
at the instant it trades therefore a policy that keeps prioritising fills at any cost (thereby being filled disproportionately on toxic flow from informed traders) profits at the
expense of the risk manager, on whom the adverse move lands through
$\Phi_t$. A blend coefficient $\lambda$ pays each agent a share of the desk total
$D_t = \psi^{b}_t + \psi^{a}_t + \Phi_t - c_t$ alongside its own term,
\begin{equation}
  r^{Q}_t = (1-\lambda)\,\tilde r^{Q}_t + \lambda\,\frac{D_t}{2},
  \qquad
  r^{R}_t = (1-\lambda)\,\tilde r^{R}_t + \lambda\,\frac{D_t}{2}.
  \label{eq:desk-blend}
\end{equation}
Such a design encourages the quoter to differentiate between the different types of flow and thus improve its overall quoting.
At $\lambda = 0$ the rewards are fully separate and the moral hazard is at its fullest. At
$\lambda = 1$  the game becomes fully common-sum and allows for little differentiation between the two roles through the reward function.
Intermediate $\lambda$ trades one
against the other. The sum is preserved at every
$\lambda$, since both agents receive $\lambda D_t/2$, so the blend
redistributes a fixed total rather than changing it.

\paragraph{Shaping} Two further terms are added to improve the shaping of the reward in \cref{eq:desk-blend}.
Firstly, a \emph{markout} which directly encourages the quoter to improve the quality of its fills by including how the mid moves 
over the steps that follow each fill, in proportion to the net amount it just traded. 
A fill followed by an adverse move is penalised even though it captured spread, so
the quoter is judged on the quality of its fills rather than exclusively on spread capture.
The second is an \emph{exit credit}: the risk manager is rewarded on
the quality of its closes by how the mid moves after the close (e.g. if it liquidates a long and the mid price increases, it is penalized). This helps further discourage constant undifferentiated liquidation.
Both terms are designed to mitigate temporal and multi-agent credit-assignment difficulties by looking forward and more directly attributing outcomes to individual responsibilities.

\paragraph{Curriculum} Finally, the strongest configuration anneals the markout terms so as to prevent excessively restrictive policy during the initial phase of learning. 
The markout coefficient is gradually increased from zero over the early part of training, so the quoter learns to quote
before it is judged on the quality of its fills. The coefficients and schedules
are given in \cref{sec:hparams}.

\newsection{Methodology}\label{sec:methodology}

\Cref{sec:background} defined our problem including the spaces, rewards, the desk and reasoning behind those design choices. With this requisite information covered, this section covers the learners trained on it, and justifies our design choices.

We retain JaxMARL-HFT's actor-critic foundation, including its PPO update rule and optimiser. Our changes predominantly lie in the problem definition as expounded in the preliminaries section. We also add an 
auxiliary prediction head (for the mid price movement). We maintain the same optimization such that improvements can be attributed to the design of the problem rather than a better-tuned learner.
We also keep the recurrent trunk of their architecture as it stands, given the gated recurrent unit (GRU) performs comparably to similar alternatives (long short-term memory (LSTM)) on these tasks while carrying significantly fewer parameters and costing less per step \citep{chung2014empirical,cho2014learning}. This decision was made in favour of 
allowing for more rapid iterations of experiments. We favour computational efficiency since our focus is once again on the problem design as a potential solution.

\subsection{Single-agent architecture}\label{sec:single}

\subsubsection{Why a policy gradient}\label{sec:pg}
Tabular learning would evidently be an intractable given the size of the observation and action spaces. As such, we first considered a value-based learner, given these methods are pervasive in the market-making literature, from the tile-coded temporal
difference learners of \citet{spooner2018market} onward.
A deep value-based learner uses an underlying neural network that takes the observation history as input and outputs an estimated
action value for each possible action, trained to match a target built from the
reward actually received plus its own estimate of the best next action's value,
the policy acts greedily on the estimates, exploring with probability
$\epsilon$. We chose not to use this approach for the following reasons:

\begin{itemize}
  \item \textbf{The deadly triad.} We want to avoid the deadly triad of
        reinforcement learning: the combination of function approximation,
        bootstrapped targets and off-policy maximisation
        \citep{sutton2018reinforcement}. A deep Q-network in this context would have all three components.
        Its target $y_t = r_t + \gamma \max_{a'} Q(s_{t+1}, a'; \theta)$
        contains the network's own estimate of the best next action, and a
        parameter update can subsequently raise that estimate for an action the policy
        never actually takes. Never visited, the overestimation is never
        corrected, its value can diverge, and the divergence spreads into
        every other estimate that uses it as part of its target. We
        therefore chose an on-policy method: targets are built from the
        actions the current policy takes, so an overestimation is corrected
        once the pair is visited again.
  \item \textbf{Catastrophic forgetting.} The observations a learner sees in this environment are partly generated by its current policy,
   thereby shifting the distribution of its experience as its policy changes. This is particularly salient with our fixed-time markets given that the optimal policy halfway through the market's life is different to the optimal policy close to resolution (where time-to-resolution dominates the value). 
   A network fitted to the most recent experiences then overwrites what it learned from earlier experiences, a term denoted as catastrophic forgetting \citep{sutton2018reinforcement}.
   Unlike policy-gradient methods, value-based learning is more exposed to this issue given that its targets are bootstrapped from a function approximator, so updates on visited observations also shift the targets of unvisited ones. Training many market windows in parallel, each at a
        different point of a different market's life, mitigates this risk for our on-policy method.
  \item \textbf{Expressiveness.} An $\epsilon$-greedy policy commits to one action per observation and spreads the remaining probability uniformly over the rest of the actions for exploration. 
  A policy-gradient method instead parametrises the policy directly and can represent any distribution over the quote
   table, This matters firstly because the optimal policy may genuinely be stochastic given the
    latent nature of order-flow toxicity. Secondly, because its exploration will be concentrated around
     its current policy carrying a lower real monetary cost in a live market relative to a random uniform exploration (from $\epsilon$-greedy).
\end{itemize}

The drawback we accept from this approach is added variance to the gradient estimates.
\subsubsection{Proximal policy optimisation with a recurrent
policy}\label{sec:ppo}

Proximal policy optimisation \citep{schulman2017ppo} is a refinement of the
actor-critic framework: an \emph{actor}, the parameterised policy, chooses
actions, and a \emph{critic} estimates the value of an observation, thereby reinforcing an action outcome based on its advantage $\hat{A}_t$ (the return obtained relative to the critic's estimate). 

A single gradient step on the policy can substantially change it and thereby degrade performance. Trust-region methods
\citep{schulman2015trpo} addressed this limitation by constraining each update to a region wherein the policy changes little, but this affected computational efficiency given a
constrained optimisation (or a penalty term) had to be applied at each step. PPO therefore refined this by replacing the
constraint with a clipped surrogate objective. Each sample is weighted by the
importance ratio
$\rho_t(\phi) = \pi(a_t \mid h_t; \phi)\,/\,\pi(a_t \mid h_t; \phi_{\text{old}})$,
of the current policy relative to the previous policy. Clipping the ratio to
$[1-\epsilon, 1+\epsilon]$ and taking the minimum of the two terms bounds how
far any update can move the policy.
\begin{equation}
\mathcal{L}(\phi_i) = \mathbb{E}\Big[
\min\big(\rho_t \hat{A}_t,\;
\operatorname{clip}(\rho_t, 1-\epsilon, 1+\epsilon)\,\hat{A}_t\big)
    \Big]
    - c_v\, \mathcal{L}^{V}(\theta_i)
    + c_e\, \mathbb{E}\big[H(\pi(\cdot \mid h_t;\,\phi_i))\big].
\label{eq:ppo}
\end{equation}
Our context is characterized by a noisy reward signal, sufficiently so that a single batch could suggest a large step in the wrong direction, making the clipping particularly important. The
critic's loss $\mathcal{L}^{V}$, its squared error to a bootstrapped target, is
clipped on the same principle. The entropy bonus $H$ discourages
collapse onto a single action which is equally important in our setting given the degenerate "No Trade" collapse.

 \paragraph{Generalised Advantage Estimation} The advantage is estimated with
generalised advantage estimation \citep{schulman2016gae}, which weights a
reward arriving $n$ steps after an action by $(\gamma\lambda)^n$. With
$\gamma$ effectively one, $\lambda$ sets how far back credit reaches, and so
directly addresses the aforementioned temporal credit assignment problem. We set to be higher than the inherited value from JaxMARL-HFT, allowing 
for credit assignment commensurate with the holding times that we measured in our markets. The exact value is given in \cref{sec:hparams}.

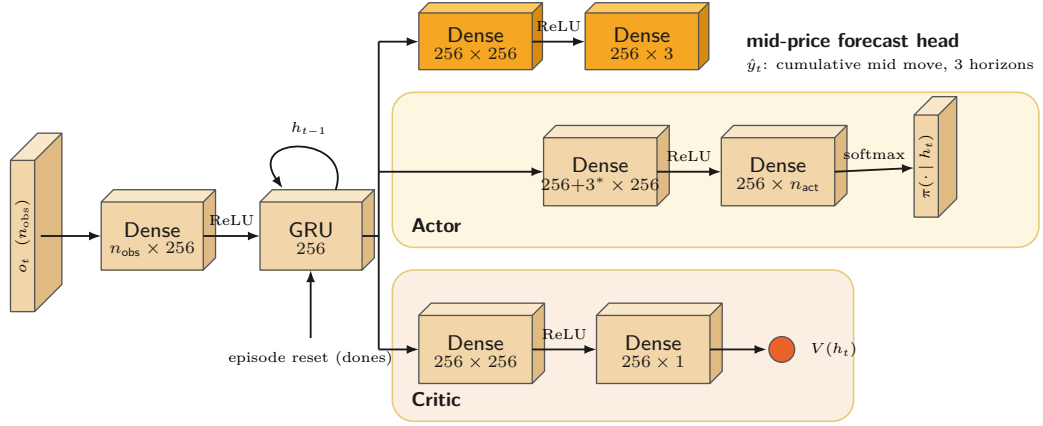
\begin{figure}[t]
\centering
\begin{tikzpicture}[font=\sffamily\color{fgink}, line join=round]

\definecolor{fgink}{HTML}{1A1A1A}
\definecolor{ctA}{HTML}{FDF6E0}\definecolor{ctB}{HTML}{FBEEE3}
\definecolor{ctBorder}{HTML}{E8C97A}
\definecolor{stdF}{HTML}{F2D5A8}\definecolor{stdT}{HTML}{F9E7C9}\definecolor{stdR}{HTML}{E0BA83}
\definecolor{empF}{HTML}{F5A623}\definecolor{empT}{HTML}{FFC85C}\definecolor{empR}{HTML}{D98B10}
\definecolor{term}{HTML}{E8622A}
\definecolor{edge}{HTML}{555555}

\newcommand{\cub}[9]{%
  \pgfmathsetmacro{\dd}{0.18*#4}%
  \fill[#6] (#1,{#2+#4}) -- ({#1+\dd},{#2+#4+\dd}) -- ({#1+#3+\dd},{#2+#4+\dd}) -- ({#1+#3},{#2+#4}) -- cycle;
  \fill[#7] ({#1+#3},#2) -- ({#1+#3+\dd},{#2+\dd}) -- ({#1+#3+\dd},{#2+#4+\dd}) -- ({#1+#3},{#2+#4}) -- cycle;
  \fill[#5] (#1,#2) rectangle ({#1+#3},{#2+#4});
  \draw[edge, line width=0.5pt] (#1,#2) rectangle ({#1+#3},{#2+#4});
  \draw[edge, line width=0.5pt] (#1,{#2+#4}) -- ({#1+\dd},{#2+#4+\dd}) -- ({#1+#3+\dd},{#2+#4+\dd}) -- ({#1+#3},{#2+#4});
  \draw[edge, line width=0.5pt] ({#1+#3+\dd},{#2+#4+\dd}) -- ({#1+#3+\dd},{#2+\dd}) -- ({#1+#3},#2);
  \node[align=center] at ({#1+#3/2},{#2+#4/2+0.09}) {\fontsize{8}{9}\selectfont #8};
  \node[align=center] at ({#1+#3/2},{#2+#4/2-0.17}) {\fontsize{6.5}{7}\selectfont #9};
}
\tikzset{arr/.style={-{Latex[length=1.6mm]}, line width=0.7pt, draw=fgink},
         ann/.style={font=\fontsize{5.5}{6}\selectfont}}

\fill[ctA, rounded corners=8pt] (5.0,0.55) rectangle (13.55,2.6);
\draw[ctBorder, line width=0.6pt, rounded corners=8pt] (5.0,0.55) rectangle (13.55,2.6);
\node[anchor=south west] at (5.12,0.62) {\fontsize{7.5}{8}\selectfont\bfseries Actor};
\fill[ctB, rounded corners=8pt] (5.0,-1.75) rectangle (11.1,0.25);
\draw[ctBorder, line width=0.6pt, rounded corners=8pt] (5.0,-1.75) rectangle (11.1,0.25);
\node[anchor=south west] at (5.12,-1.68) {\fontsize{7.5}{8}\selectfont\bfseries Critic};

\cub{-0.05}{-0.3}{0.32}{2.0}{stdF}{stdT}{stdR}{}{}
\node[ann, rotate=90] at (0.11,0.7) {$o_t$\; ($n_{\text{obs}}$)};
\cub{1.15}{0.25}{1.35}{0.9}{stdF}{stdT}{stdR}{Dense}{$n_{\text{obs}}\times 256$}
\cub{3.25}{0.25}{1.35}{0.9}{stdF}{stdT}{stdR}{GRU}{$256$}
\draw[arr] (0.35,0.7) -- node[ann, above]{} (1.15,0.7);
\draw[arr] (2.5,0.7) -- node[ann, above]{ReLU} (3.25,0.7);
\draw[arr] (4.25,1.31) .. controls (4.5,2.05) and (3.05,2.05) .. (3.55,1.33);
\node[ann] at (3.9,2.12) {$h_{t-1}$};
\draw[arr] (3.92,-0.65) -- (3.92,0.25);
\node[ann, anchor=north] at (3.92,-0.68) {episode reset (dones)};

\draw[line width=0.7pt, draw=fgink] (4.6,0.7) -- (4.82,0.7) -- (4.82,1.55);
\draw[line width=0.7pt, draw=fgink] (4.82,0.7) -- (4.82,-0.8);
\draw[arr] (4.82,1.55) -- (7.0,1.55);
\draw[arr] (4.82,-0.8) -- (5.35,-0.8);

\draw[line width=0.7pt, draw=fgink] (4.82,1.55) -- (4.82,3.27);
\draw[arr] (4.82,3.27) -- (5.35,3.27);
\cub{5.35}{2.9}{1.5}{0.75}{empF}{empT}{empR}{Dense}{$256\times 256$}
\cub{7.55}{2.9}{1.5}{0.75}{empF}{empT}{empR}{Dense}{$256\times 3$}
\draw[arr] (6.85,3.27) -- node[ann, above]{ReLU} (7.55,3.27);
\node[anchor=west] at (9.55,3.27)
  {\fontsize{7.5}{8}\selectfont\bfseries mid-price forecast head};
\node[ann, anchor=west] at (9.55,2.95) {$\hat y_t$: cumulative mid move, 3 horizons};

\cub{7.0}{1.1}{1.5}{0.9}{stdF}{stdT}{stdR}{Dense}{$256{+}3^{\ast}\times 256$}
\cub{9.35}{1.1}{1.5}{0.9}{stdF}{stdT}{stdR}{Dense}{$256\times n_{\text{act}}$}
\draw[arr] (8.5,1.55) -- node[ann, above]{ReLU} (9.35,1.55);
\cub{11.9}{0.95}{0.32}{1.35}{stdF}{stdT}{stdR}{}{}
\node[ann, rotate=90] at (12.06,1.62) {$\pi(\cdot \mid h_t)$};
\draw[arr] (10.85,1.55) -- node[ann, above]{softmax} (11.9,1.6);

\cub{5.35}{-1.25}{1.5}{0.9}{stdF}{stdT}{stdR}{Dense}{$256\times 256$}
\cub{7.7}{-1.25}{1.5}{0.9}{stdF}{stdT}{stdR}{Dense}{$256\times 1$}
\draw[arr] (6.85,-0.8) -- node[ann, above]{ReLU} (7.7,-0.8);
\fill[term] (10.15,-0.8) circle (0.17);
\draw[edge, line width=0.5pt] (10.15,-0.8) circle (0.17);
\draw[arr] (9.2,-0.8) -- (9.95,-0.8);
\node[ann, anchor=west] at (10.4,-0.8) {$V(h_t)$};

\end{tikzpicture}
\caption[The network]{Our network differs from \citet{frey2023jaxlob}
in three respects: the layers are $256$ wide rather than $128$, given the significantly larger observation. 
The actor's head is a softmax over the quote table (instead of multivariate normal) since the action space is
discrete. A third head outputs the mid's cumulative move,
in ticks, one, five and ten steps ahead (\cref{sec:hparams}), trained by
squared error against the realised move, weighted at $0.1$ in \cref{eq:ppo},
dropping windows that cross a rollout/episode boundary rather than zeroing them. In some configurations noted in \cref{sec:hparams} its forecast
is also appended to the actor's input under a stop-gradient to encourage the policy to act on the forecast. The $+3^{\ast}$ term is present only in those configurations; otherwise the actor receives the $256$-dimensional recurrent representation. The stop-gradient ensures that the head remains trained on its own target. $n_{\text{obs}}$
and $n_{\text{act}}$ vary by seat as given in \cref{sec:obs} and \cref{sec:actions}.}
\label{fig:network}
\end{figure}

\paragraph{The network.}\label{sec:network} The architecture follows the RNN-PPO network from
\citet{frey2023jaxlob} illustrated in \cref{fig:network}. This recurrence is
required given \cref{sec:decision-model} established that the state is not fully
observed. The third head in \cref{fig:network} forecasts the mid price at
several step horizons. It is designed to give the agent visibility on where the mid is going and thereby inform their quoting and inventory management. In the multi-agent context, it is suited to 
complement the quoter and the risk managers' markout and exit
credit terms respectively, since both these terms (\cref{sec:reward}) judge the agent by where
the mid goes after it trades. The head is an auxiliary task inspired by \citet{jaderberg2017reinforcement}. 
Its prediction error is backpropagated into the shared layers such that the
embedding and GRU learn to encode the information that is predictive of mid price movement (which is fed into the actor and the critic). 
The observation (\cref{sec:obs}) already contains order imbalance, the feature most predictive of short-term mid price movements \citep{cont2014price}, so this head
serves simply to ensure that the recurrent unit learns to use this information rather than discard it. In the
configurations noted in \cref{sec:hparams} the forecast is also appended to the
actor's input, so that the policy can condition on it directly.

\subsection{Multi-agent proximal policy optimisation}\label{sec:marl}

\paragraph{Independent learners.} We use independent proximal policy optimisation (IPPO) to train our agents. This method
allows for a simple extension of single-agent algorithms to a multi-agent setting, namely by running one copy of the algorithm per 
agent and letting them treat each other as part of their environment. This approach is a stronger baseline in cooperative settings than its simplicity suggests
\citep{dewitt2020independent}. Concretely, each agent type has its own network
and its own optimiser. 

\paragraph{Formulation.} Agent $i$ has policy $\pi(\cdot\,;\phi_i)$ and
critic $V(\cdot\,;\theta_i)$. Both terms are conditioned on $h^i_t$, the recurrent
state of \cref{sec:ppo} summarising agent $i$'s observations so far (i.e. at time $t$) in the
episode. Under independent learning each agent runs single-agent PPO on
its own trajectory: the gradient is taken with respect to $\phi_i$ alone,
\begin{equation}
\nabla_{\phi_i} J(\phi_i) \;\propto\;
\mathbb{E}\Big[\,
    \hat{A}_i\big(h^i_t, a_i\big)\,
\nabla_{\phi_i} \log \pi\big(a_i \mid h^i_t;\, \phi_i\big)
  \Big],
\label{eq:ippo}
\end{equation}
Where $\hat{A}_i$ denotes the GAE of \cref{sec:ppo} (computed from agent $i$'s own
rewards and critic). This is an
approximation of the multi-agent policy gradient theorem
\citep{lowe2017maddpg, foerster2018counterfactual, kuba2021settling,
lyu2023centralized}, whose exact form credits the joint action of all agents
given their joint history, which no agent observes given this formulation.

\subsubsection{Independent versus central learning}\label{sec:multiagent}
A central learner instead treats all agents as a single decision maker choosing a
joint action. It largely circumvents the challenges of non-stationarity and credit assignment that arise in independent learning. 
We rejected it given that the joint action space would be the product of the individual action spaces of the quoter and the risk manager, 
which would become computationally intractable with our stated action spaces in \cref{sec:actions-desk}. Independence keeps each agent's action
space rich (\cref{sec:quote-table}) and lets the design scale, at two costs.

\begin{itemize}
\item \textbf{Non-stationarity.} A central learner controls a joint policy and thereby does not face
  the specific non-stationarity issue faced by the independent learner. Namely, an independent learner's environment contains the
  other agents, whose policies change as they learn. This can produce cyclic co-adaptation whereby an agent changes its policy
  to adjust for the other agent's policy, and in turn the other agents change their policies in response, and so on. This 
  therefore removes any convergence guarantee and could lead to instability during learning. 
  We opt not to correct for it given that JaxMARL-HFT reports convergence for market-making and
  execution agents in this setting \citep{mohl2025jaxmarlhft}. Although real markets are themselves non-stationary, the source here is simultaneous policy adaptation and should not be interpreted as a faithful model of market adaptation. The
  remaining risk is that agents co-adapt to a few learned participants rather than the whole order flow; rectifying this is left for future work.

\item \textbf{Credit assignment.} Temporal credit assignment is already
  hard for one agent as has been widely addressed in the previous
  sections. Multiple agents add the question of whose actions earned the reward
  \citep{foerster2018counterfactual}. A central learner can condition value estimates on the joint action, allowing outcomes associated with one agent's action to be distinguished across different actions of its partner and thereby providing richer information for multi-agent credit assignment. Independent learners do not explicitly represent these joint-action values (as described in \cref{sec:reward-desk}),
  and this also motivated the general-sum split of \cref{sec:reward-desk} as well as the isolated rewards of the markout and exit credit.
  Independent learners could be given communication actions \citep{foerster2016learning} to explicitly model collaboration, but these
  would widen the action space and add a further layer of credit assignment given that
  an agent must infer the meaning of a received message and learn which message to send, which would 
  require a further layer of reward design complexity (to ensure rewarded communication actions are disentangled from rewards earned through the agent's own decisions).
  Instead each agent observes a block describing its partner
  (\cref{sec:obs-desk}) and infers coordination from what the partner's
  actions did. The environment is our implicit communication channel as previously stated.
\end{itemize}

The market maker shares its book with a learned execution agent that optimizes the execution of a scheduled order (inherited from \citep{mohl2025jaxmarlhft}), 
amounting to a mixed-motive game. We chose to keep this because an adapting learner is a more demanding opponent than replayed historical flow
or a zero-intelligence agent and thereby may robustify our agents' policies.

\newsection{Experimental Setup}\label{sec:setup}
This section instantiates the problem described in \cref{sec:background} and the method in
\cref{sec:methodology} as a concrete experiment.

\subsection{Market assumptions}\label{sec:assumptions}

We build on JaxMARL-HFT and inherit its market assumptions along with their subsequent
limitations. This results in the following limitations acknowledged below:
\begin{itemize}
  \item \textbf{No transaction costs.} No fee is charged for trading which is at odds with the venue's published rules. This was chosen in order to keep the scope of the research tractable.
  \item \textbf{Non-reactive flow.} While the matching engine matches historical orders against the quotes of the agent(s), the recorded market participants do not behaviorally respond to the agents' actions. Specifically, the endogenous behavioural response to another market participants (the agents) is absent in the historical messages: no market participant
  is re-pricing, withdrawing or changing its behaviour as a function of what our agents quote. This is tantamount to the assumption of negligible price impact in the market which simplifies our simulation.  The execution agent partly
  remedies this, since it is the one participant that does continually adapts to the
  agents' quotes.
    \item \textbf{The reconstructed book's queue composition is an assumption of
        the data}, not something the data records. This is expounded upon in
        \cref{sec:data}.
\end{itemize}

\subsection{The simulation environment}\label{sec:env}

\Cref{sec:sim} described the simulator, this section fixes the quantities our
experiments run it at, collected in \cref{tab:envfixed}.

\begin{figure}[htbp]
\centering
\includegraphics[width=\textwidth]{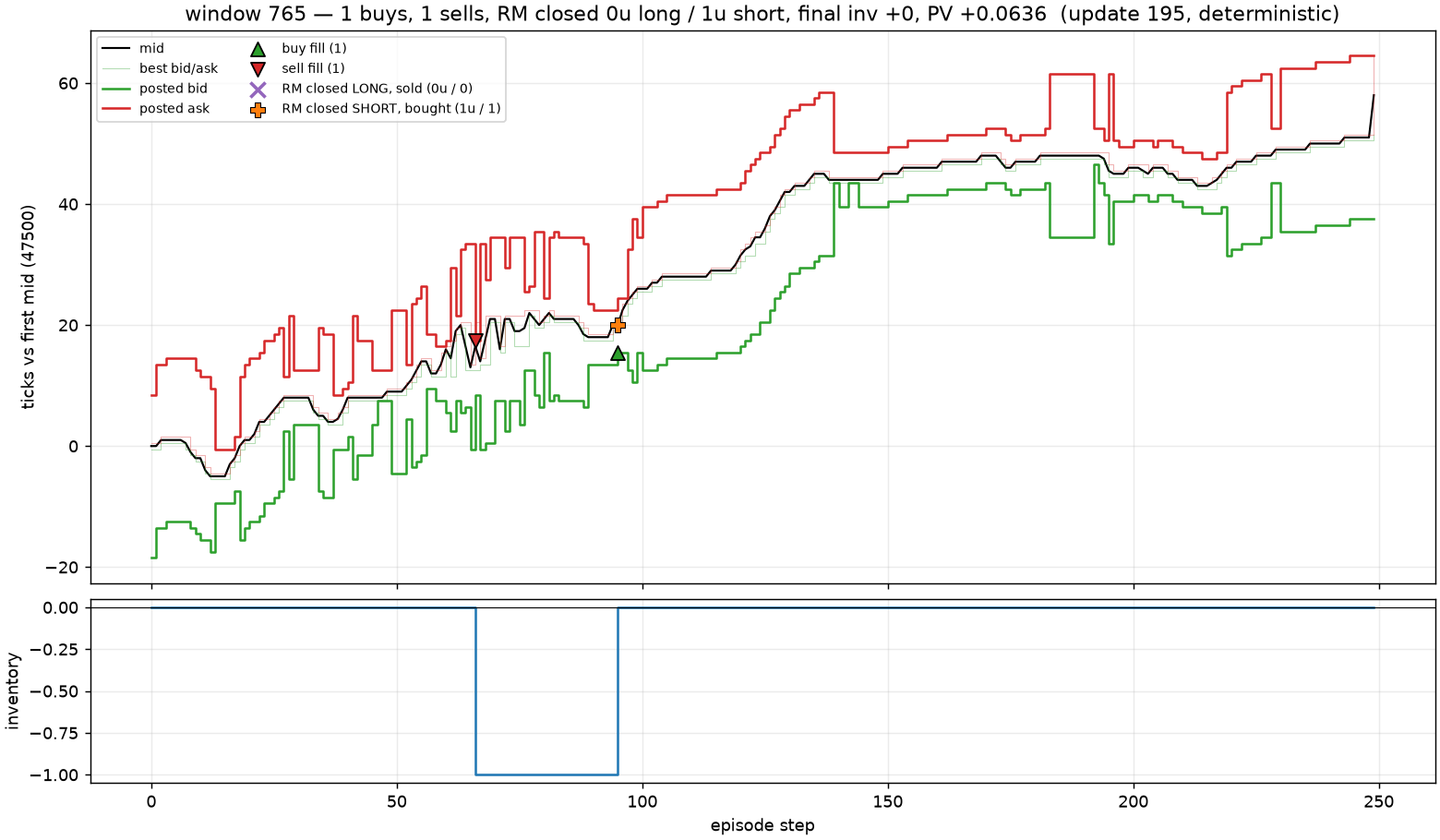}
\caption[A held-out episode traded by the desk]{One held-out episode (window 765 of the 1{,}152 held-out windows),
traded by a \texttt{D\_MARK} checkpoint. Prices are in ticks relative to the
window's first mid; the black line is the mid, the thin lines the venue's
best bid and ask, and the green/red steps the desk's posted bid and ask. A
sell fill opens a one-unit short (lower panel: inventory), which the risk
manager later buys back, ending the episode flat. PV is the episode's
end-of-episode portfolio value in ticks.}
\label{fig:quotes}
\end{figure}

An episode corresponds to the replay of one five-minute market from its recorded messages, ending
at the markets resolution. Each step advances 400 recorded messages, and an episode is
250 steps, so the stretch of recorded messages it replays (referred to as its window), is 100{,}000 messages long. The agents' own messages are
inserted among these and do not count toward the window. Given that a step is a
fixed number of messages rather than a fixed duration, the market time an
episode covers depends on how message-dense the market is. A typical held-out episode
spans roughly its final two minutes. Training runs 64 parallel
environments (independent copies of the simulator that take simultaneous steps and pool their transitions into a single update). 
Each environment then draws its next window at random from the pool of training markets. Held-out markets are never seen in training (to prevent leakage). 
\Cref{fig:quotes} shows one held-out episode as the desk trades it.

\begin{table}[htbp]
\centering
\caption{Quantities fixed across every configuration.}
\label{tab:envfixed}
\small
\begin{tabular}{ll}
\toprule
quantity & value \\
\midrule
recorded messages per step   & 400 \\
episode length               & 250 steps (100{,}000 messages) \\
book depth                   & 10 levels a side loaded, 5 observed \\
parallel environments        & 64 \\
transitions per update       & $64 \times 250$, in 2 minibatches, 4 epochs \\
configured budget            & 3.2 million environment steps (200 updates) \\
hardware                     & one NVIDIA RTX 4090 GPU, one AMD Ryzen 7 9800x3d \\
data                         & the markets and split of \cref{sec:data} \\
\bottomrule
\end{tabular}
\end{table}

The observation, action and reward spaces, and the division of labour inside
the desk, are as defined in \cref{sec:background,sec:methodology}.

\subsection{Data}\label{sec:data}

\subsubsection{Market choice and split}\label{sec:data-market}

\begin{figure}[htbp]
\centering
\includegraphics{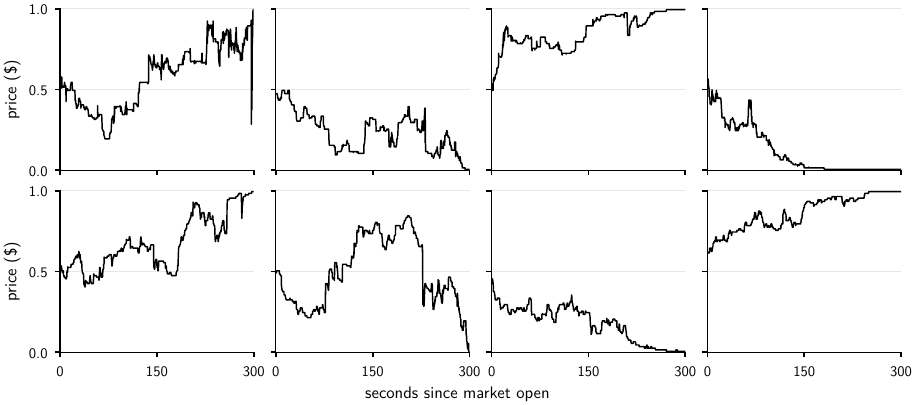}
\caption[Eight of the five-minute markets]{Eight of the five-minute markets: one from each training day (top
row, in date order) and four from the held-out day (bottom row). Each panel is
the Up token's mid price from open to settlement.}
\label{fig:data-panel}
\end{figure}

\textbf{The instrument.} We trade Polymarket's
\href{https://polymarket.com/event/btc-updown-5m-1787837100}{BTC up/down
five-minute}
binary markets. A new market opens every five minutes, resolves at a deadline
fixed when it opens, and pays exactly 0 or 1. Each market compares the price
of Bitcoin across five minutes and bets are placed on whether the final price is
higher or lower than a reference price (BTC's last-traded price set five minutes before resolution),
making it precisely a cash-or-nothing digital option on Bitcoin. While markets open
earlier than five minutes before resolution, we restrict ourselves to the
final five minutes given that this is when the reference price becomes known and the
market gains volume and liquidity. Prices lie between \$0.01 and \$0.99. We settled on this instrument
after first working on a deeply liquid and
\href{https://polymarket.com/event/fed-decision-in-july-181}{slow-paced
federal rate prediction} which had a much longer time-horizon and may have been influenced by institutional participants hedging risk. 
We favor our choice of market since firstly exposing our agent to the full breadth of the market up until resolution allows us to train it
on hundreds of different markets and thereby against hundreds of different endogenous price formation scenarios, giving it a more comprehensive view of possible situations in markets.
Secondly, the liquidity provision on these particular markets is thin and thus motivates the need for servicing these markets.
\Cref{fig:data-panel} shows eight of these markets.

\textbf{Source and split.} We capture eight hours of markets a day, from 12:00
to 20:00 UTC. Training uses four days (22 and 28 April, 14 and 28 May 2026). The held-out set is one whole day two weeks later, 11 June 2026. 
We split by whole day since markets five minutes apart
share the same Bitcoin path and the same participants, so a market whose
neighbour is in the other set may not be a sufficiently representative evaluation test. 

\textbf{Tokens.} A market trades through two tokens, Up and Down, and we use
both. Each market yields two message streams, one per token. This gives
640 training and 192 held-out streams and balances the settled outcomes,
with half settling at 1. 

\subsubsection{From book updates to order messages}\label{sec:data-conversion}

The simulator requires market-by-order (MBO) level data (every individual order carrying an identifier).
Polymarket only publishes and collects L2-orderbook data (the total resting size at each price level). 
We decided to remain within this framework and infer MBO data from the L2 data. Since the exchange itself only publishes and stores L2 historical orderbook data (and no MBO),
this tradeoff was appropriate given the matching engine's realism and the speed with which it runs on the hardware.

\textbf{Reconstruction.} The raw L2 data (the full L2 order-book at every order-book update stamped in milliseconds) was obtained from
\href{https://www.pmxt.dev}{PMXT}, a data provider for prediction
markets. In order to validate the integrity of their data, although its archive has already been used in published research \citep{nechepurenko2026perps}, we separately validated its historical data against Polymarket's own API websockets and found the two to be consistent.
From timestamped L2 data one can infer an order arrival when a
level's total size rises, and an order departure when the size falls, while an accompanying trade
print allows us to differentiate the order departures between cancels and fills. Since the simulator requires the individual 
orders behind each total (each message is an order), we synthesize them using this inference detailed above.
Each observed increase becomes one order message carrying the observed time, price and size, so the
output stream is the recorded feed re-encoded as orders, and the
reconstructed totals equal the reported totals throughout. Only the order
identities are invented, and they matter for queue position.
Price-time priority fixes two of the three update rules exactly: an arrival
joins the back of its queue, and a fill consumes the front. The one
assumption is the cancellation: which order (in the queue) is cancelled is not in the data 
and with nothing to favour one position we remove one drawn uniformly at
random. This is a limitation of the work, though not a departure from the
simulator's own practice: JaxLOB initialises each window from an L2 snapshot
by treating every level as a single order of the summed volume
\citep{frey2023jaxlob}, the same kind of assumption applied to the opening
book rather than to every cancellation. A more faithful reconstruction is left to future work. The output is
written as LOBSTER message and order-book file pairs \citep{huang2011lobster}
with the message vector of \cref{eq:message}, commensurate with the requirements of the \citep{frey2023jaxlob} engine.

\subsection{Models}\label{sec:models}
Six model configurations are compared: one baseline, one non-desk model of
ours, and four desk configurations. Our models share the stack of \cref{tab:stack} and
each is named for what it changes, thereby allowing every comparison to isolate a single design choice.Every configuration, the baseline included, trains alongside a learned
execution agent quoting into the same book.

\begin{table}[htbp]
\centering
\caption{The stack shared by our models.}
\label{tab:stack}
\small
\begin{tabular}{ll}
\toprule
component & setting \\
\midrule
quote table     & the seventeen-row wide table of \cref{sec:quote-table}, plus stand-aside \\
observation     & the book-and-flow vector of \cref{sec:obs} \\
auxiliary head  & the mid-price forecast head of \cref{sec:network}, coefficient $0.1$ \\
inventory cap   & 4 units \\
network         & 256 recurrent units, 256 per dense layer \\
\bottomrule
\end{tabular}
\end{table}

\textbf{Baseline}
\begin{itemize}
  \item \texttt{PAPER\_BASE}: the JaxMARL-HFT market maker reproduced
        verbatim and retrained on our data: its nine-row quote table, its
        two-number observation (spread and inventory), its asymmetrically
        damped reward with a quadratic inventory penalty (this defines the best-performing configuration from their paper
        \citep{mohl2025jaxmarlhft}), as well as an inventory
        cap of twenty units.
\end{itemize}

\textbf{Non-desk models.}
\begin{itemize}
  \item \texttt{WIDE\_SPOONER}: our full stack with the richer observation and action spaces. Under the reference's
        reward, so its comparison with \texttt{PAPER\_BASE} is controlled for
        the objective.
\end{itemize}

\textbf{Desk models.} Each is the split desk of \cref{sec:reward-desk} plus
one change, except the last, which combines the two repairs.
\begin{itemize}
  \item \texttt{D\_SPLIT}: quoter and risk manager under the general-sum
        split detailed in \cref{sec:reward-desk}.
  \item \texttt{D\_BLEND}: the split with a blend that moves a share of
        each agent's reward toward the desk total.
  \item \texttt{D\_MARK}: the split with the quoter additionally rewarded a markout on its fills as in \cref{sec:reward-desk}.
  \item \texttt{D\_BL\_MARK\_CURR}: both blend and markout alongside the training curriculum of \cref{sec:reward-desk}, at the settings given in \cref{sec:hparams}.
\end{itemize}

\subsection{Experiment settings}\label{sec:hparams}
\Cref{tab:ppo} gives the optimisation settings shared by every model.

\begin{table}[htbp]
\centering
\caption[Settings common to every model]{Settings common to every model. Where a value differs by agent
type, both are given (i.e. market maker, execution). The quoter and risk manager take the market maker's
value unless otherwise specified.}
\label{tab:ppo}
\small
\begin{tabular}{ll}
\toprule
setting & value \\
\midrule
optimiser                        & Adam, learning rate $2.5\times10^{-4}$, constant \\
discount $\gamma$                & $\approx 1$ (market making, risk manager); $0.999$ (execution) \\
GAE trace $\lambda$              & $0.97$ (market making, risk manager); $0.9$ (execution) \\
clipping $\epsilon$              & $0.2$ \\
entropy coefficient              & $0.01$ \\
value coefficient                & $0.5$ \\
maximum gradient norm            & $0.5$ \\
recurrent width / layer width    & $256$ / $256$, ReLU \\
update epochs / minibatches      & $4$ / $2$ \\
environments $\times$ steps      & $64 \times 250$ per update \\
total environment steps          & $3\,200\,000$, i.e.\ $200$ updates \\
held-out evaluation              & every fifth update, a full $250$-step rollout \\
\bottomrule
\end{tabular}
\end{table}
\paragraph{Hyperparameters.} At $\lambda = 0.97$ the effective credit horizon
of the GAE trace is roughly thirty-three steps, longer than the holding times
we measure (\cref{sec:ppo}); the cost is variance, since a longer horizon sums
more noise into every estimate. This and ten other hyperparameters were
searched with Bayesian optimisation via Weights \& Biases sweeps: the three
loss-stop settings, the blend, markout and forecast coefficients, the learning
rate, entropy coefficient, update epochs and inventory cap. Trials ran at
reduced scale, 32 environments for 60 updates, so fractional annealing schedules were defined relative to the duration of
each run.

\paragraph{Forecast head.} Every model of ours carries the mid-price forecast
head at coefficient $0.1$; \cref{tab:aux} gives the horizons and whether the
forecast is appended to the actor's input. The head is switched per run, not
per agent, so every learner in a run carries it.

\begin{table}[htbp]
\centering
\caption{Forecast-head settings.}
\label{tab:aux}
\small
\begin{tabular}{lll}
\toprule
model & horizons (steps) & forecast to actor \\
\midrule
\texttt{D\_SPLIT}, \texttt{D\_BLEND}, \texttt{D\_MARK} & 1, 5, 10 & no \\
\texttt{WIDE\_SPOONER} & 1, 5, 10 & yes \\
\texttt{D\_BL\_MARK\_CURR} & 1, 10, 60 & yes \\
\bottomrule
\end{tabular}
\end{table}

\paragraph{Loss stop.} A loss stop halts the opening of new risk for the rest
of a window once the episode's loss passes a threshold in ticks
(\cref{sec:reward-enhancements}), annealed from
a first value to a second over a fraction of the run. \texttt{WIDE\_SPOONER} runs $120 \to 30$ over the first $30\%$;
\texttt{D\_BL\_MARK\_CURR} runs $240 \to 15$ over the first $40\%$. No other
model carries one.  

\paragraph{Desk reward terms.} \Cref{tab:desk} gives the blend and markout per
desk model. Every desk model's risk manager additionally receives the exit
credit of \cref{sec:reward-desk} at $0.01$ over a ten-step horizon.

\begin{table}[htbp]
\centering
\caption[Desk alignment terms]{Desk alignment terms. Markout is given as a coefficient (that is applied to the markout term in the quoter's reward), where it is valid, it applies with a step horizon of $5$.
Ramped means the coefficient is scaled in linearly over the first $30\%$ of the run.}
\label{tab:desk}
\small
\begin{tabular}{lll}
\toprule
model & blend & markout \\
\midrule
\texttt{D\_SPLIT}          & $0$   & -- \\
\texttt{D\_BLEND}          & $0.3$ & -- \\
\texttt{D\_MARK}           & $0$   & $0.3$ / $5$ \\
\texttt{D\_BL\_MARK\_CURR} & $0.3$ & $0.15$ / $5$, ramped \\
\bottomrule
\end{tabular}
\end{table}

Each model was run once at a fixed seed, so the dispersions reported in
\cref{sec:results} are across held-out windows, not seeds. The models build on the public \href{https://github.com/vmohl/JaxMARL-HFT}{JaxMARL-HFT} Github repository,
with our modifications clearly specified in our previous sections. Full training configurations and implementations are available upon request.

\subsection{Evaluation criteria}\label{sec:eval}
Given the differing reward functions and configurations across models, we cannot directly compare the performance of the models directly against their rewards. Instead,
we opt for evaluating each model on a proxy metric that reflects the performance of the market-maker at the end of each market:
\textbf{end-of-episode portfolio value}, the cash a model holds once an
episode ends and any remaining position is closed. 
JaxMARL-HFT report average end-of-episode portfolio value as a training curve. We include this,
and additionally score our models on mean end-of-episode portfolio value measured on held-out windows (i.e. our validation windows) to 
compare the performance of our models against unseen data.

\begin{itemize}
\item \textbf{Level and settling.} We evaluate held-out portfolio value
  every fifth update over 64 held-out windows. We report the mean of the
  last five evaluations as the model's conclusive level, and the standard deviation of
  the last ten serves to indicate the stability of the training process.
  Training curves are included where the learning process itself is of
  interest.
\item \textbf{No-trade share.} Portfolio value is trivially improved by not
  trading as previously described. We therefore
  report alongside every portfolio value the share of steps on which the
  model selects the no-trade action, and treat an improvement accompanied
  by a large rise in that share as withdrawal from trading rather than successful learning.
\item \textbf{Reward decomposition.} Portfolio value is separated into
  spread capture and profit on held inventory. This directly reflects whether
  a model's quoting or its carry determines its result.
\item \textbf{Distribution across windows.} For our desk models, we additionally evaluate their final
  policy deterministically on held-out windows and report the resulting distribution of end-of-episode portfolio value. This allows
  for a more granular evaluation of the performance of our design choices.
\end{itemize}

These methods collectively give us an understanding of the consequences of our design choices expounded in the previous
sections. Each model is run once at a fixed seed, so every dispersion is across windows rather than seeds
(\cref{sec:limitations}). 

\newsection{Results}\label{sec:results}
We report two comparisons, each testing the design choices that motivated them.
The first covers single market making agents (non-desk), against the most successful published
architecture in \citep{mohl2025jaxmarlhft}. We seek firstly to determine whether
the richer observation and action spaces of \cref{sec:obs,sec:actions} are
tractable, improve learning,
and whether, combine \cref{sec:obs,sec:actions} with controlling inventory, (outside the reward) this prevents the
no-trade collapse. The second covers the desk configurations and asks whether the reward
design of \cref{sec:reward-desk} resolves the multi-agent credit-assignment problem that the
split creates and whether independent learning remains stable (despite non-stationarity) in practice.
Every desk model inherits the single-book stack of \cref{tab:stack}.
\subsection{Non-desk models}\label{sec:results-single}

Two configurations trade the same held-out windows under the same reward.
\texttt{PAPER\_BASE} is the IPPO-RNN market maker of JaxMARL-HFT retrained on
our data. \texttt{WIDE\_SPOONER} keeps its reward and replaces its
observation and action spaces with the stack of \cref{tab:stack}, so the
comparison holds the reward objective fixed while evaluating the changes to observations, actions, and inventory controls.

\begin{figure}[htbp]
\centering
{\sffamily\small held-out windows\par}
\includegraphics[width=\textwidth]{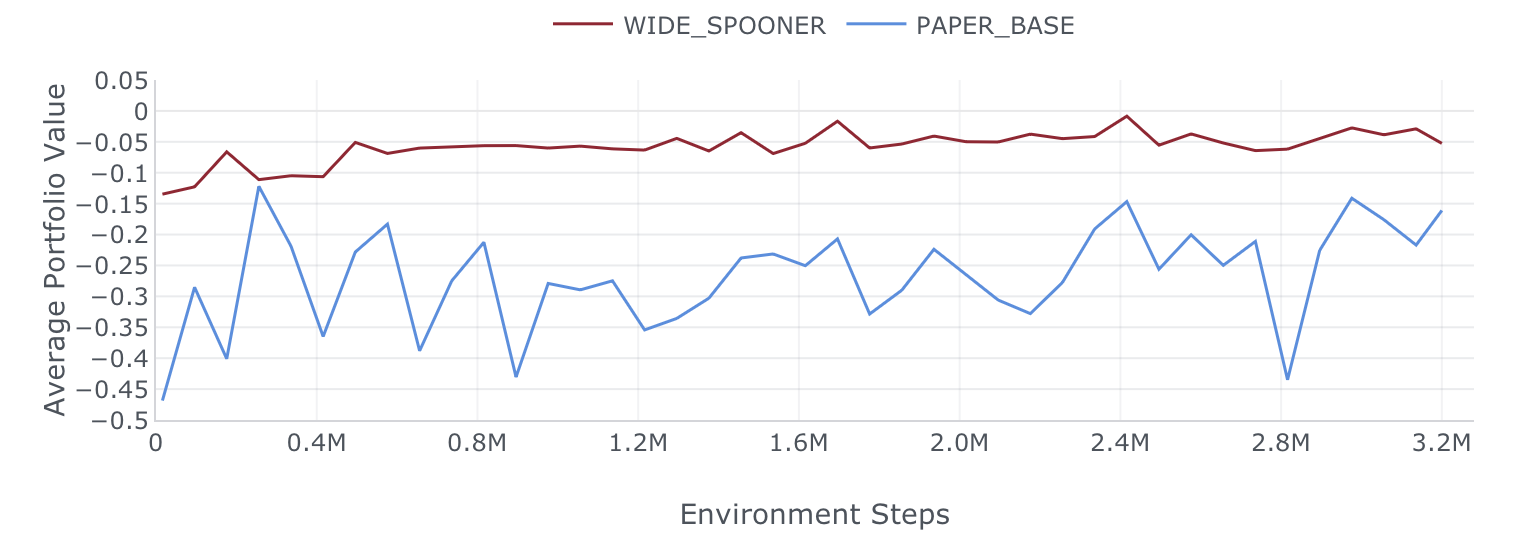}\\[6pt]
{\sffamily\small training windows\par}
\includegraphics[width=\textwidth]{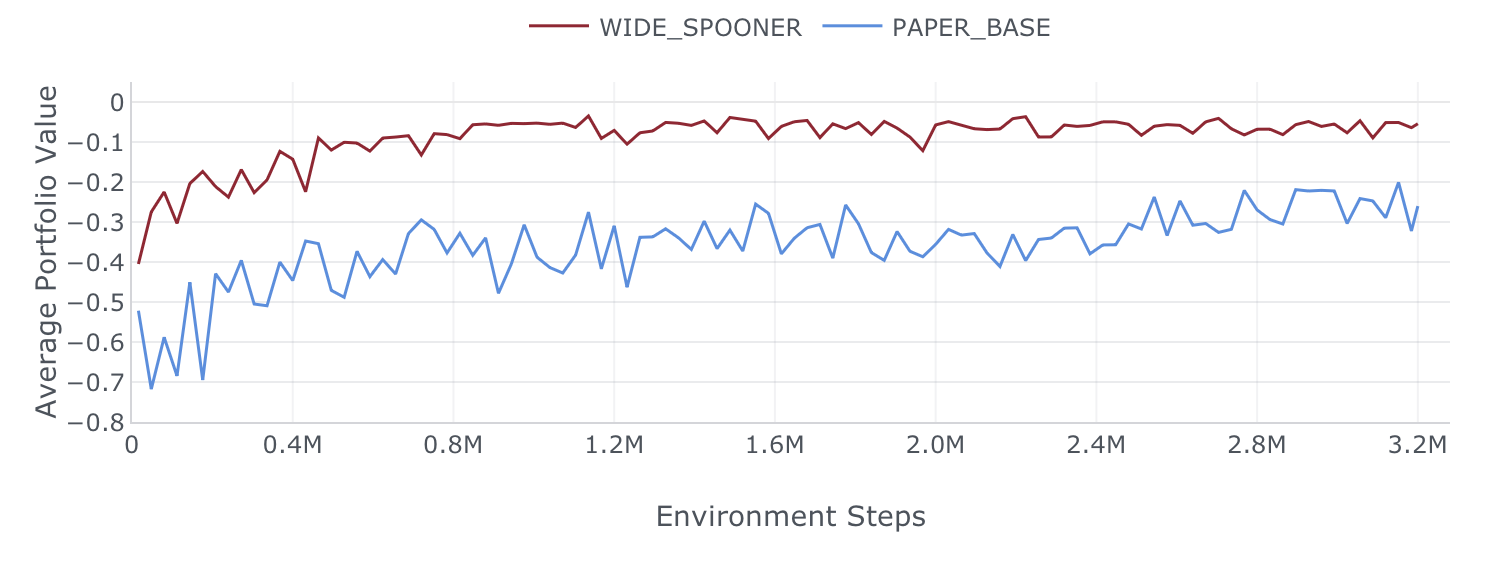}
\caption[Single-book training curves]{End-of-episode portfolio value during training for
\texttt{WIDE\_SPOONER} and \texttt{PAPER\_BASE}, on held-out windows (top,
evaluated every fifth update) and training windows (bottom, every update).
One seed per model.}
\label{fig:single-book}
\end{figure}

Clearly our model outperforms the baseline, finishing at $-9.6$ ticks
per episode against $-46.1$ (\cref{fig:single-book}, \cref{tab:nondesk}) in the held-out windows. In \cref{fig:single-book}, the bottom 
panel shows that the variance of the portfolio value throughout training is less volatile in our model
than that of the paper baseline. While this could naively be attributed to the inventory cap and the loss stop, we note that
\texttt{WIDE\_SPOONER} rises from $-101$ to $-23$ ticks within the
first $0.8$M steps (the majority of its eventual gain) and settles near
$-16$, while \texttt{PAPER\_BASE} rises from $-130$ to $-64$ over the full
budget and stays noisy. The learned gain is thus larger for our model ($85$ ticks, against $66$) and clearly learns faster, which could not be explained by the inventory cap alone. 
Notably, this is corroborated by the held-out evaluation, which 
demonstrates our model generalizing to held-out unseen market windows with far more stability than
the baseline and steadily improving throughout training. In contrast, while the baseline policy 
seems to be converging in training, its evaluation on unseen data is consistent with the richer specification supporting better generalization, thus putting into question the quality of its learned policy.

Since the reward is shared, the held-out performance difference is consistent with the redesigned configuration supporting better generalisation. However, the observation and action spaces change alongside inventory controls, so these experiments cannot identify which component accounts for the difference or establish how much additional market structure was learned.

Another factor that would contradict our analysis is the no-trade collapse, since a higher portfolio value attained through withdrawal 
from trading would not be a meaningful improvement. However, \texttt{WIDE\_SPOONER} selects the no-trade action on 4.4\% of steps during training while
the baseline picks this action 19.2\%, which elucidates that the improvement does not come at the expense of direct withdrawal from trading.
The redesigned configuration therefore trades more frequently while losing less on these held-out windows. A likely mechanism beyond the improved learning from the observation space could be the action space:
\texttt{PAPER\_BASE}'s nine symmetric rows quote both sides evenly or not at
all, whereas ours admits a less restrictive possible space of actions. Together with the aforementioned reasons,
this empirically validates the design choices of \cref{sec:obs,sec:actions} and the stack of \cref{tab:stack}.

\begin{figure}[htbp]
\centering
\includegraphics{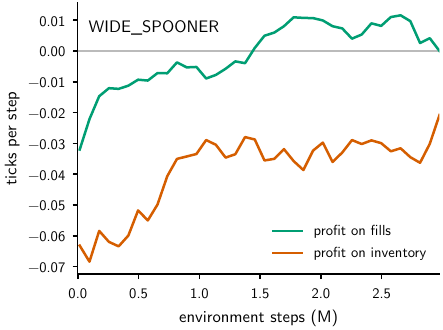}
\caption[The two parts of the reward]{The two parts of \texttt{WIDE\_SPOONER}'s reward during training,
under the same marking convention. Spread capture turns positive, the
inventory term stays negative and is the larger of the two throughout, so it
decides the total.}
\label{fig:reward-terms}
\end{figure}

Portfolio value nonetheless remains negative, and the plot in \cref{fig:reward-terms}
illustrates the possible reasoning behind this. The realized PnL (previously also referred to as spread capture)
turns positive (per step i.e. per decision) for our model during training, but is disproportionately dampened by the increasing
but still negative unrealized PnL (previously also referred to as inventory mark-to-market) term.
It is this measurement that motivated the desk formulation of the problem by separating these two underlying components of the reward.

\subsection{Desk models}\label{sec:results-desk}
Our comparison tests two claims. \Cref{sec:reward-desk} argued that the
bare split would cause a multi-agent credit assignment issue leading to moral hazard, and that the blend would partly restore this alignment,
while the markout would explicitly restore it for the quoting agent. We further
further seek to verify whether the non-stationarity of independent learning (with co-adapting agents) 
would cause unstable learning under our design.

\begin{figure}[htbp]
\centering
{\sffamily\small held-out windows\par}
\includegraphics[width=\textwidth]{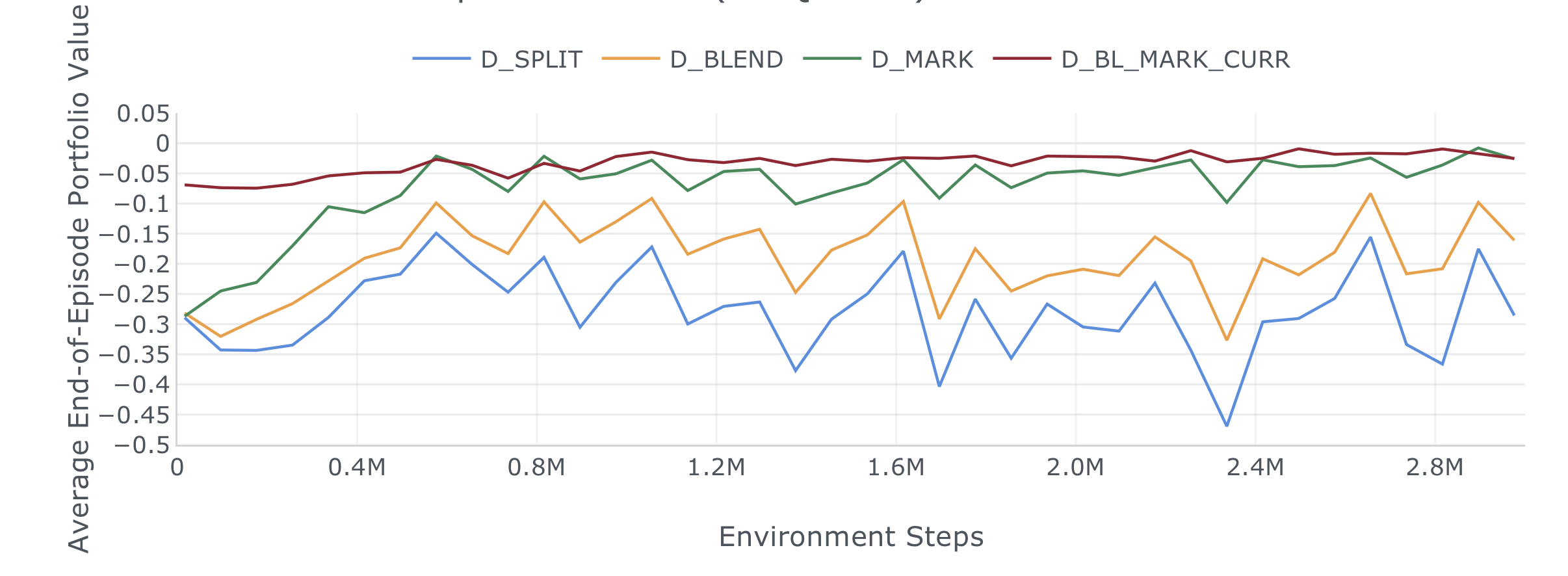}\\[6pt]
{\sffamily\small training windows\par}
\includegraphics[width=\textwidth]{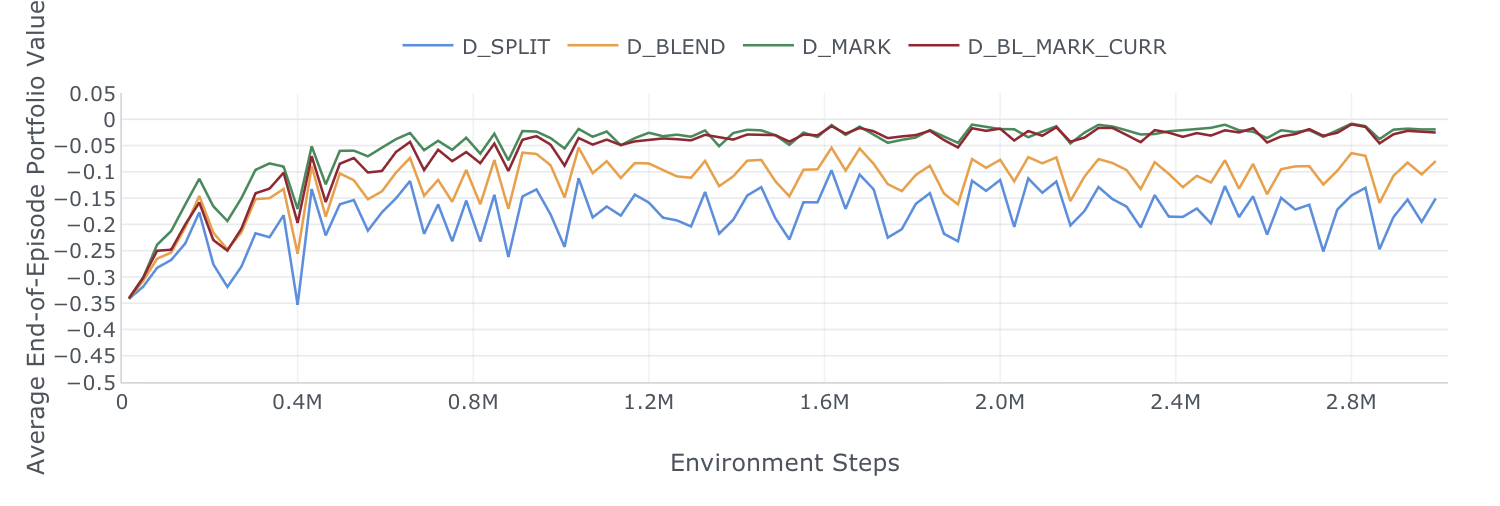}
\caption[Desk training curves]{End-of-episode portfolio value during training for the desk
configurations of \cref{tab:desk}, on held-out windows (top, evaluated
every fifth update) and training windows (bottom, every update). One seed
per model.}
\label{fig:desk}
\end{figure}

\textbf{Comparing the training performance of the desk configurations.} 
\texttt{D\_SPLIT} finishes at $-72.3$ ticks per episode, below
\texttt{PAPER\_BASE} and the worst configuration of our paper. This can partly be attributed
to the multi-agent credit-assignment problem since the quoter is rewarded for spread capture
while the risk manager is penalized for the inherited inventory, thus not allowing
the quoter to differentiate between temporally profitable and unprofitable fills.
\Cref{fig:desk} provides preliminary evidence of this being learned. After
approximately 0.8M environment steps, held-out end-of-episode PV exhibits a negative trend for both \texttt{D\_SPLIT} and \texttt{D\_BLEND}. Between
0.816M and the environment step endpoint, PV falls from $-47.3$ to $-71.5$ for \texttt{D\_SPLIT} and from $-24.3$ to $-40.3$ for \texttt{D\_BLEND}, 
despite continued improvement in their training objectives. For \texttt{D\_SPLIT}, this
deterioration coincides with a 21\% increase in mean absolute inventory
and a 27\% increase in traded quantity relative to the 0.4-0.8M-step interval. 
These observations would be consistent with the split objective encouraging excessive trading and inventory exposure that does not generalize out of sample.
In the bottom panel we see that \texttt{D\_BLEND} performs comparatively better than the alternative \texttt{D\_SPLIT}.
Since a part of the desk's result now returns
to the quoter, which lifts the level but does not stop the decay in held-out PV nor the variance exhibited in the training PV. 
\texttt{D\_MARK}'s performance in contrast demonstrates steeper learning in the training process, and contains a policy that generalizes in the held-out set. This is tangible evidence
that the markout term improves the quality of the agents' fills in this configuration. The price move after a fill enters the objective of
the quoter that took the fill, which adds the missing information without
diluting either reward. This is the ordering we predicted \cref{sec:reward-desk} and clearly
demonstrates that the markout term is the most effective of the two reward modifications.

\textbf{The curriculum results in the best performance.}
\texttt{D\_BL\_MARK\_CURR} finishes its held-out evaluation at a PV of $-4.3$ ticks, the best in the paper, at
the smallest evaluation spread of any configuration we ran ($1.7$ against
\texttt{D\_SPLIT}'s $17.6$). It also starts at $-17$ where the other
three start near $-72$, because its loss stop additionally bounds what an untrained
policy can lose from the first update. Measured as the gain from start to
finish, \texttt{D\_MARK} learns more. What the combination visibly adds is
stability, not level.

\textbf{Independent learning did not destabilise the desk.} The
configurations with careful reward design clearly lead to stable learning. The two that degrade do so
only in the held-out set and under a reward we have argued is misaligned. With one seed per
configuration this is an observation, not a guarantee.

\begin{figure}[htbp]
\centering
\includegraphics{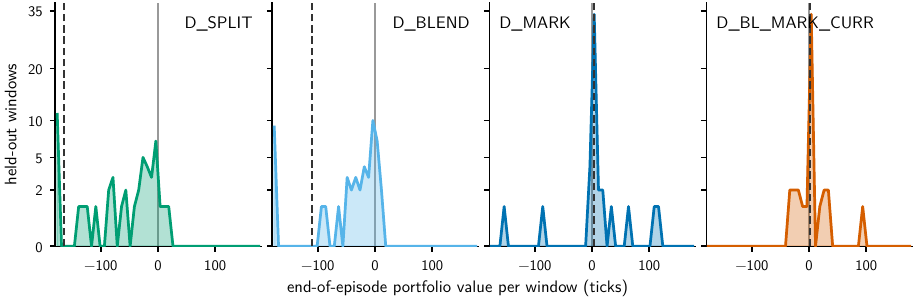}
\caption[Held-out results window by window]{Held-out results window by window, one panel per desk
configuration, identical axes, 48 windows each. The dashed line marks each
distribution's mean; windows beyond the axis are shown at its edge, ten,
eight, zero and zero from left to right.}
\label{fig:histograms}
\end{figure}

\Cref{fig:histograms} shows the same four configurations window by window.
\texttt{D\_SPLIT} loses in 45 of 48 windows and \texttt{D\_BLEND} in 39,
with ten and eight windows respectively beyond the axis. \texttt{D\_MARK}
loses in 6 and \texttt{D\_BL\_MARK\_CURR} in 8, both finish positive in
aggregate ($+133$ and $+67$ ticks over the 48), and neither has a window
beyond the axis. The repairs collapse a wide, left-heavy distribution onto
zero. This is the tail-risk channel the shaping terms were designed to
close, and it is closed by the markout: \texttt{D\_MARK}, which carries no
curriculum, removes as many losing windows as the combination, so the
curriculum cannot be credited with the tail on this evidence. Its visible
contribution is the narrower spread around zero in the rightmost panel.

\subsection{Performance comparison}\label{sec:results-overall}

\begin{table}[htbp]
\centering
\caption[Non-desk results]{Non-desk configurations. Portfolio value is the held-out
end-of-episode value in ticks, averaged over the last five
evaluations; spread is its standard deviation over the last ten, so a smaller
figure is a run that has settled. No-trade is the share of training steps on
which the policy selects the no-trade action. One seed per configuration.}
\label{tab:nondesk}
\small
\setlength{\tabcolsep}{5pt}
\begin{tabular}{lrrrrr}
\toprule
configuration & PV & spread & volume & no-trade & $|$inv$|$ \\
 & (ticks) & & (units/step) & (\%) & (units) \\
\midrule
PAPER\_BASE & $-46.1$ & $19.4$ & $0.056$ & $19.2$ & $1.108$ \\
WIDE\_SPOONER & $-9.6$ & $3.1$ & $0.032$ & $4.4$ & $0.366$ \\
\bottomrule
\end{tabular}

\end{table}

\begin{table}[htbp]
\centering
\caption[Desk results]{Desk configurations, same columns. The middle two rows each add one
term to \texttt{D\_SPLIT}; the last combines both with the loss-stop
curriculum.}
\label{tab:desk-results}
\small
\setlength{\tabcolsep}{5pt}
\begin{tabular}{lrrrrr}
\toprule
configuration & PV & spread & volume & no-trade & $|$inv$|$ \\
 & (ticks) & & (units/step) & (\%) & (units) \\
\midrule
D\_SPLIT & $-72.3$ & $17.6$ & $0.112$ & $1.6$ & $0.241$ \\
D\_BLEND & $-41.0$ & $12.3$ & $0.068$ & $4.7$ & $0.172$ \\
D\_MARK & $-5.9$ & $3.5$ & $0.032$ & $20.4$ & $0.082$ \\
D\_BL\_MARK\_CURR & $-4.3$ & $1.7$ & $0.022$ & $19.7$ & $0.060$ \\
\bottomrule
\end{tabular}

\end{table}

Read against the design, the results fall as follows. The wider observation
and action spaces are tractable, and the harder learning problem we accepted
for them turned out faster and steadier than the baseline's: the wide
configuration reaches its level in a quarter of the update budget and
settles at a sixth of the baseline's spread. The redesigned configuration has a lower no-trade share, but the comparison does not isolate whether inventory control outside the reward caused this difference. Independent learning did not prevent the adequately
rewarded desk from settling. The bare split turns that into
moral hazard, which the blend partly and the markout largely repair, in the
order the reward design predicted. The repairs act on the tail of the
distribution rather than its mean, and the curriculum contributes stability
rather than level.

\subsection{Limitations}\label{sec:limitations}
The shortcomings of the method and of the data are the following:

\textbf{The counterparty.} Every learned configuration trains against a
learned execution agent, which makes the environment non-stationary from the
market maker's perspective and means the comparisons hold under that setting
rather than against replayed flow alone. A control without the execution agent
was staged but cut from the scope of the paper.

\textbf{The data.} One instrument family on one venue, four training days and
a held-out day two weeks later. While our holdout is far later and prevents leakage, the breadth across instruments, regimes and venues is untested.

\textbf{The evaluation.} Inference is over held-out windows, not over training
seeds: every reported configuration is a single seed. Since Deep RL is notoriously seed-sensitive \citep{henderson2018deep}, this is noted
as a limitation and a possible avenue of future work. 

\textbf{The reconstructed book.} Although inferred order-level messages reproduce reported aggregate depth at each update, order identities and queue positions are synthetic. Fill dynamics therefore depend on the reconstruction assumptions and cannot be equated with observed historical execution.

\newsection{Conclusion}\label{sec:conclusion}
Our study brings multi-agent reinforcement learning to market making
on prediction markets. On extracted Polymarket data, we refined \citep{mohl2025jaxmarlhft}'s single-book market maker with book-and-flow observations, 
a less restrictive action space with explicit no-trade and one-sided actions, and various reward designs.
We further introduced a cooperative decomposition of the market maker problem into two seats: a quoter and a risk manager, designing appropriate
rewards to mitigate multi-agent credit assignment challenges. Every configuration was scored
on held-out end-of-episode portfolio value and other metrics, evaluated against the published market maker \citep{mohl2025jaxmarlhft}
which was retrained on the same data at a matched update budget. Our
results support four insights:

\begin{itemize}
\item \textbf{The redesigned single-seat configuration improves held-out results.} \texttt{WIDE\_SPOONER} finishes at $-9.6$ ticks per
  episode against $-46.1$ for \texttt{PAPER\_BASE}, reaches its level in a
  quarter of the budget, and settles at a sixth of the baseline's spread.
  The gain is not bought by withdrawal: it selects the no-trade action on
  $4.4\%$ of steps against the baseline's $19.2\%$, so controlling inventory
outside the reward prevents the no-trade collapse without paying for
inaction.
\item \textbf{The desk decomposition is successful when properly designed.} The bare split is the worst configuration in the
  paper at $-72.3$ ticks, below the baseline, because the quoter is paid for
  fills whose losses land on the risk manager. The blend partly repairs this
  ($-41.0$), the markout largely ($-5.9$), and their combination under the
  curriculum gives the best result we have ($-4.3$), in the order the reward
  design predicted.
\item \textbf{The repairs act disproportionately on the tail.} Losing held-out
  windows fall from 45 of 48 under the bare split and 39 under the blend to
  6 and 8 under the markout configurations, which finish positive in
  aggregate ($+133$ and $+67$ ticks over the 48). These deterministic-window aggregates do not establish positive expected profitability, and all configurations retain negative mean portfolio value under periodic held-out evaluation. The curriculum does not
  reduce the tail further. Its contribution is stability.
\item \textbf{Independent learning did not destabilise the desk.} Every
  adequately rewarded configuration converges, and the better
  configurations are also the settled ones: spread falls from $19.4$ to
  $3.1$ ticks across the non-desk comparison and from $17.6$ to $1.7$ across
  the desk ladder.
\end{itemize}

This study is built to be continued, and several directions follow directly
from it:

\begin{itemize}
  \item \textbf{Cross-market collaboration.} The venue lists many markets
        whose prices are logically constrained by one another as mentioned in the related
        work section. A collaborative desk quoting several linked books at once, leveraging 
        the design seen in this implementation, is an interesting avenue of research.
  \item \textbf{Richer multi-agent system} Giving risk manager access to other correlated instruments to hedge the risk of the market making agent.
  \item \textbf{Training Foundation Model.} Training a foundation model using the abundance of available Polymarket
        data together with the GPU-accelerated framework. This would allow us thereafter to investigate possible fine-tuning strategies for specific markets or types of markets. 
  \item \textbf{Policy self-play and adversarial curricula.} Leveraging policy self-play in order to investigate whether this results in more robust, generalizable policies and faster learning
        would complement well with the previous two points.
\end{itemize}

Each of these is a study of its own, and the environment, data pipeline and
evaluation protocol built here are the ground they would stand on.

\clearpage
\renewcommand{\refname}{Bibliography}
\bibliography{references}
\addcontentsline{toc}{section}{Bibliography}
\end{document}